%% file: iclr2027_conference.tex
\documentclass{article} 
\usepackage{iclr2027_conference,times}

\input{math_commands.tex}

\usepackage{xcolor}
\definecolor{darkblue}{RGB}{0,0,139}
\usepackage{hyperref}
\hypersetup{
    colorlinks=true,
    citecolor=darkblue,
    linkcolor=darkblue,
    urlcolor=darkblue
}
\usepackage{url}

\usepackage{xspace}
\newcommand{\benchname}{BEHAVIOR-Spatial\xspace}   

\newcommand{\vqaname}{BEHAVIOR-Spatial-VQA\xspace}
\newcommand{\taskvqaname}{BEHAVIOR-Spatial-TaskVQA\xspace}
\newcommand{\IRname}{BEHAVIOR-Spatial-IR\xspace}
\newcommand{\datasetname}{BEHAVIOR-Spatial-Action\xspace}
\newcommand{\vlaname}{BEHAVIOR-Spatial-VLA\xspace}

\usepackage{array}
\newcolumntype{L}[1]{>{\raggedright\arraybackslash}p{#1}}
\usepackage{graphicx}
\usepackage{float}
\usepackage{booktabs}

\title{%
  \parbox{\textwidth}{%
    \centering
    \fontsize{16}{19}\selectfont
    \mbox{BEHAVIOR-Spatial Benchmark:}\\
    \mbox{Evaluating Spatial Understanding for Robotics}\par
    \vspace{10pt}
  }%
}

\author{%
  \parbox{\dimexpr\textwidth-2\tabcolsep\relax}{%
    \centering
    Sirui Chen$^{1,*}$, Huang Huang$^{1,*}$,
    Qineng Wang$^{2}$, Wenlong Huang$^{1}$, \\
    Manling Li$^{2}$, Ruohan Zhang$^{1}$,
    Li Fei-Fei$^{1}$, Jiajun Wu$^{1}$ \\[2pt]
    {\normalfont
      $^{1}$Stanford University \quad
      $^{2}$Northwestern University
    } \\[2pt]
    {\normalfont\small $^{*}$Equal contribution.}
  }%
}

\usepackage{wrapfig}
\begin{document}

\newif\ifarxiv
\arxivtrue        

\ifarxiv\iclrfinalcopy\fi

\maketitle
\ifarxiv
  \fancyhead{}
  \renewcommand{\headrulewidth}{0pt}
\fi

\begin{abstract}
Robot tasks require grounding spatial language in physical scenes, but task success alone cannot establish whether spatial instructions are followed. We present \benchname, a benchmark built on BEHAVIOR-1K that measures spatial question answering, point-based target grounding, and robot action to examine how VLM spatial understanding relates to VLA behavior. It comprises 3,256 spatial questions across 49 activities, 1,240 grounding queries across 14 manipulation tasks, and action evaluations covering those 14 tasks and 10 navigation tasks. Our manipulation protocol evaluates held-out configurations and instruction following when the requested target changes within a fixed training layout. The baseline achieves 41.6\% on held-out configurations but only 16.1\% under changed instructions. 
VLM adaptation and robot-data pretraining provide limited improvements, whereas oracle visual target cues raise these rates to 60.3\% and 38.0\%, respectively. Perception evaluations reveal policy-dependent correlations between VLM question-answering and grounding accuracy and VLA manipulation performance. Together, these findings motivate evaluating spatial instruction following alongside task completion and examining when spatial perception performance translates into robot behavior. Project website: \url{https://arielchen07.github.io/behavior-spatial/}.
\end{abstract}

\input{fig_text/fig_teaser}

\input{text/sec1_intro}

\input{text/sec2_related_work}

\input{text/sec3_benchmark}

\input{text/sec4_vla_model}



\input{text/sec6_experiments_results}

\input{text/sec7_limitations}



\subsection*{AI use statement}

Generative AI tools were used to assist with drafting and editing
manuscript text, writing figure-generation scripts,
and aggregating evaluation results from logged metrics. The authors reviewed the AI-assisted text and code and verified the reported results against the experimental logs. The authors take responsibility for the final manuscript, including all claims and artifacts.



\subsection*{Ethics statement}


Our benchmark uses simulated environments and programmatically generated labels. The grounding failures identified in this work highlight potential risks of deploying robots that do not reliably follow instructions. Simulation performance should not be interpreted as evidence of safe real-world deployment.


\subsection*{Reproducibility statement}


To support reproducibility, we describe benchmark construction and label generation in
Section~\ref{sec:data-curation}, evaluation splits and protocols in
Section~\ref{sec:eval-protocol}, and experimental settings in
Section~\ref{sec:experiments}. We will release the benchmark
and code upon publication.




\bibliography{iclr2027_conference}
\bibliographystyle{iclr2027_conference}

\appendix
\input{supp}

\end{document}

%% file: math_commands.tex
\usepackage{amsmath,amsfonts,bm}

\def\eqref#1{equation~\ref{#1}}

\def\1{\bm{1}}

\DeclareMathAlphabet{\mathsfit}{\encodingdefault}{\sfdefault}{m}{sl}
\SetMathAlphabet{\mathsfit}{bold}{\encodingdefault}{\sfdefault}{bx}{n}



%% file: fig_text/fig_teaser.tex
\begin{figure*}[ht]
    \centering
    \includegraphics[width=0.9\textwidth]{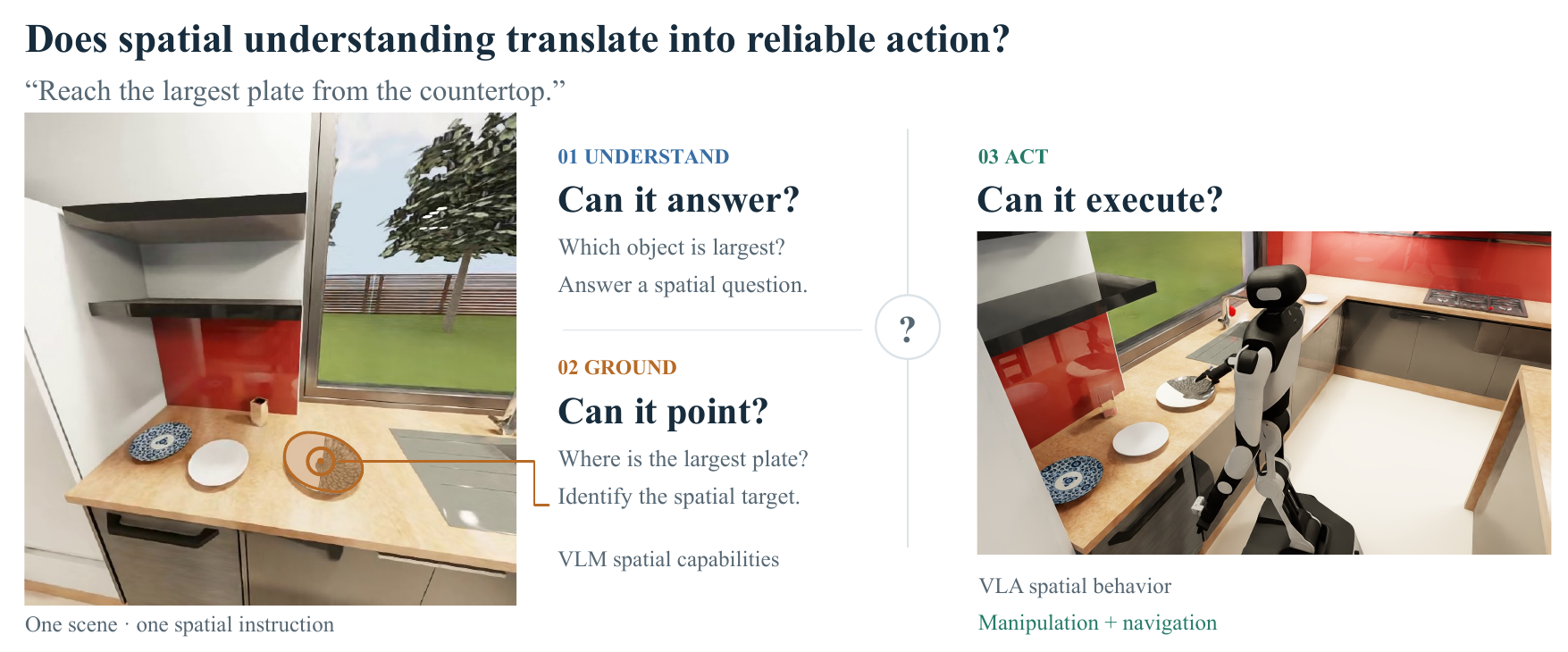}
    \vspace{-1em}
    \caption{\textbf{BEHAVIOR-Spatial} links VQA (Understand), target grounding (Ground), and execution (Act), testing whether VLM spatial capabilities predict VLA task performance.}
    \vspace{-1em}
    \label{fig:teaser}
\end{figure*}

%% file: text/sec1_intro.tex
\section{Introduction}
\label{sec:intro}

Instructions such as ``reach the second tool from the left'' require a robot to resolve a spatial relation, identify the intended target, and execute an action. Navigation similarly requires interpreting destinations relative to current and previously observed surroundings. Vision-language-action models (VLAs) are initialized from vision-language models (VLMs) and assume that the backbone's spatial understanding supports robot behavior, yet this connection is rarely studied directly. Task success alone provides ambiguous evidence: a policy may exploit familiar layouts without following spatial language, or select the correct target but fail during execution. Relating VLM spatial capabilities to VLA behavior requires jointly evaluating spatial perception, target grounding, and action.

We introduce \benchname, a benchmark built on BEHAVIOR-1K ~\citep{li2024behavior1k} that connects these evaluations across household manipulation and navigation. It comprises three components: (1)~\vqaname, a spatial VQA suite containing 3,256 questions across 49 task activities, including 1,572 general spatial questions and 1,684 task-specific questions, with both single- and multi-frame evaluations; (2)~\IRname, a point-based intermediate representation (IR) generation suite containing 1,240 target-grounding queries across 14 manipulation tasks; and (3)~\datasetname, a robot action suite covering the same 14 manipulation tasks and 10 navigation tasks. Shared scenes and relations across suites enable studying how VLM question-answering and grounding accuracy relate to VLA performance.

Our evaluation distinguishes generalization to new configurations from responsiveness to spatial instructions. \emph{Held-out} manipulation and navigation configurations test generalization, while a manipulation \emph{prompt probe} changes the requested target within a training layout to test whether behavior follows the instruction. We measure VQA answer accuracy and whether generated points fall within the instructed target region. For robot action, we report \emph{soft success} as a behavioral proxy for target-selection intention, alongside \emph{wrong-target commitments} and \emph{strict success}.\looseness=-1

Our experiments reveal persistent spatial instruction-following failures, especially when the target changes within a familiar layout. The tested LoRA adaptation and DROID~\citep{khazatsky2025droidlargescaleinthewildrobot} initialization strategies provide limited improvements on this probe, whereas supplying IR visual target cues substantially improves target-consistent behavior. These cues isolate the benefit of providing the grounding decision, rather than demonstrating learned grounding from language. For two of our best manipulation variants, the dot-IR and attention-map policies, grounding accuracy correlates more strongly with soft success than VQA accuracy does. However, the perception–action relationship varies across policy variants, and direction-question accuracy shows little correlation with navigation success across the ten evaluated tasks.\looseness=-1

Our contributions are threefold: (1)~\textbf{a spatial benchmark} connecting question answering, point-based target grounding, and household manipulation and navigation; (2)~\textbf{a diagnostic evaluation protocol} that tests changed instructions in training layouts and separates target-consistent behavior from task completion; and (3)~\textbf{an empirical study} of policy adaptation and visual grounding interfaces relating VLM spatial understanding to VLA behavior.

%% file: text/sec2_related_work.tex
\section{Related Work}
\label{sec:related}






\paragraph{Spatial understanding in VLMs and its benchmarks.}
A growing body of work studies spatial reasoning in vision-language
models, including 3D-grounded spatial QA~\citep{chen2024spatialvlmendowingvisionlanguagemodels,cheng2024spatialrgptgroundedspatialreasoning},
relative position and multi-view correspondence
(What'sUp~\citep{kamath2023whatsupvisionlanguagemodels}, BLINK~\citep{fu2024blinkmultimodallargelanguage}), egocentric video
(VSI-Bench~\citep{yang2025thinkingspacemultimodallarge}), spatial mental models
(MindCube~\citep{wang2026mindcubespatialmentalmodeling}), and atomic embodied capabilities such as pointing
(BEAR~\citep{qi2026dissectingembodiedabilitiesmultimodal}). Robo2VLM~\citep{chen2025robo2vlmvisualquestionanswering} is closest to our VQA setting,
generating questions from real robot trajectories. These benchmarks largely assess spatial question answering or pointing in isolation. \benchname instead aligns VQA,
pointing, and execution on the same scenes,
objects, and relations, testing whether spatial understanding transfers to grounded action.\looseness=-1

\paragraph{VLMs as the foundation of VLA models.}
Vision-language-action models augment pretrained VLM backbones with action prediction or control modules~\citep{brohan2023rt2visionlanguageactionmodelstransfer, kim2024openvlaopensourcevisionlanguageactionmodel, octomodelteam2024octoopensourcegeneralistrobot, black2026pi0visionlanguageactionflowmodel, intelligence2025pi05visionlanguageactionmodelopenworld, gr00tn1_2025, gemini2025robotics, hao2025mimoembodiedxembodiedfoundationmodel},
implicitly relying on the backbone's semantic and spatial knowledge
to transfer into the policy. Recent work has begun to examine this
assumption: VLM4VLA~\citep{zhang2026vlm4vlarevisitingvisionlanguagemodelsvisionlanguageaction} studies backbone choice;
\cite{hancock2025actionslanguagefinetuningvlms} identify catastrophic forgetting of
vision-language abilities during action finetuning; and Qwen-VLA~\citep{qwenvla}
uses joint vision-language and action training to mitigate this issue.
OpenHelix~\citep{cui2025openhelixshortsurveyempirical} and Mind-to-Hand~\citep{tang2025mindhandpurposefulrobotic} further incorporate
explicit embodied reasoning. Our benchmark complements these efforts
by separately measuring VQA, referent grounding, and execution.\looseness=-1

\paragraph{Benchmarks and evaluation for robot policies.}
CALVIN~\citep{mees2022calvinbenchmarklanguageconditionedpolicy}, LIBERO~\citep{liu2023liberobenchmarkingknowledgetransfer}, VLABench~\citep{zhang2024vlabench}, and
SimplerEnv~\citep{li24simpler} evaluate language-conditioned policies primarily
through end-task success, while BEHAVIOR-1K~\citep{li2024behavior1k} provides the
long-horizon household activities and simulator we build on. Factored evaluations such as The Colosseum~\citep{pumacay2024colosseumbenchmarkevaluatinggeneralization} and DOMINO~\citep{fang2026generalizableroboticmanipulationdynamic}
expose specific sources of policy brittleness. \benchname similarly
adopts a diagnostic evaluation, but targets spatial language
grounding: prompt-probe pairs hold the training scene fixed while varying the
instruction, and outcome buckets and strict/soft metrics separate
grounding from execution.\looseness=-1

\paragraph{Intermediate representations and explicit grounding for policies.}
Points, marks, traces, and sketches have been used as interfaces between perception and control~\citep{molmoact2025, li2026robointerholisticintermediaterepresentation, zhang2026spatialactionsgroundingvisionlanguageaction}.
We use such an intermediate representation for a different purpose: as a controlled intervention on an otherwise fixed policy. Providing an object-center mark tests whether supplying the correct referent resolves the grounding failure and isolates the remaining execution gap. We then ask whether the grounding signal can instead
be supplied by a VLM at rollout time.\looseness=-1

%% file: text/sec3_benchmark.tex
\section{The Behavior-Spatial Benchmark}
\label{headings}

\input{fig_text/fig_dataset_showcase}
\subsection{Design Overview}
\label{sec:design-overview}

\benchname comprises three components, overviewed in Figure~\ref{fig:dataset_showcase}.

\vqaname evaluates spatial understanding through general
and task-grounded questions. The general subset covers
spatial relations within individual views and reasoning
across multiple views, while the task-grounded subset
queries objects and relations relevant to the robot tasks.
\IRname evaluates whether a model can ground a spatial
instruction by predicting a point within the specified
target object or placement region. Finally, \datasetname
evaluates spatially grounded action through manipulation
and navigation tasks in which spatial instructions determine
the intended target or destination.

Task-grounded VQA and \IRname{} (together, the task-grounded perception suites) share task contexts, objects, and spatial relations with the action suite, enabling task-level comparison between VLM perception and VLA behavior.

\subsection{Spatial Taxonomy}
\label{sec:taxonomy}

\benchname covers direction, distance, and size.
Direction includes relative-position relations in
object-, and robot-relative frames, as well
as region identification (\emph{locate}) and spatial
ordering (\emph{ordinal}). Distance and size require
comparisons of physical separation and object size,
respectively. Spatial VQA further distinguishes
single-frame understanding from multi-frame reasoning
across robot rotation or navigation.
Appendix~\ref{app:spatial-taxonomy} details the concepts and settings.

\input{fig_text/fig_data_curation}

\subsection{Dataset Construction}
\label{sec:data-curation}

We construct \benchname from BEHAVIOR-1K demonstrations,
using recorded observations and actions together with
object geometry and camera metadata recovered through
OmniGibson replay as needed.
Figure~\ref{fig:curation_pipeline} summarizes the
pipeline. We curate \textbf{14 manipulation tasks}
and \textbf{10 navigation tasks} across five navigation
scenes. Manipulation tasks require identifying a target
among spatially distinct candidates; navigation tasks
require grounding an instructed destination.
Instruction variants specify different target objects
or goal regions, and curated scene-context traversals
support navigation.
Appendix~\ref{app:task-suite} describes task curation
and lists the tasks and variants.

A metadata-to-template pipeline generates spatial
questions and their answers, with visibility checks,
relation-margin filters, and sampling controls to
reduce ambiguity and imbalance. General VQA spans
49 activities and contains 1,078 single-frame and
494 multi-frame questions. Task-grounded VQA derives
from the curated robot task instances and contains
1,082 manipulation and 602 navigation questions.
For point grounding, we compute target-object
references and placement positions or regions from
segmentation and simulator geometry.
VQA and grounding use distinct visibility criteria.
See Appendix~\ref{app:data-curation} for curation details.

%% file: fig_text/fig_dataset_showcase.tex
\begin{figure*}[t!]
    \centering
    \includegraphics[width=0.98\textwidth]{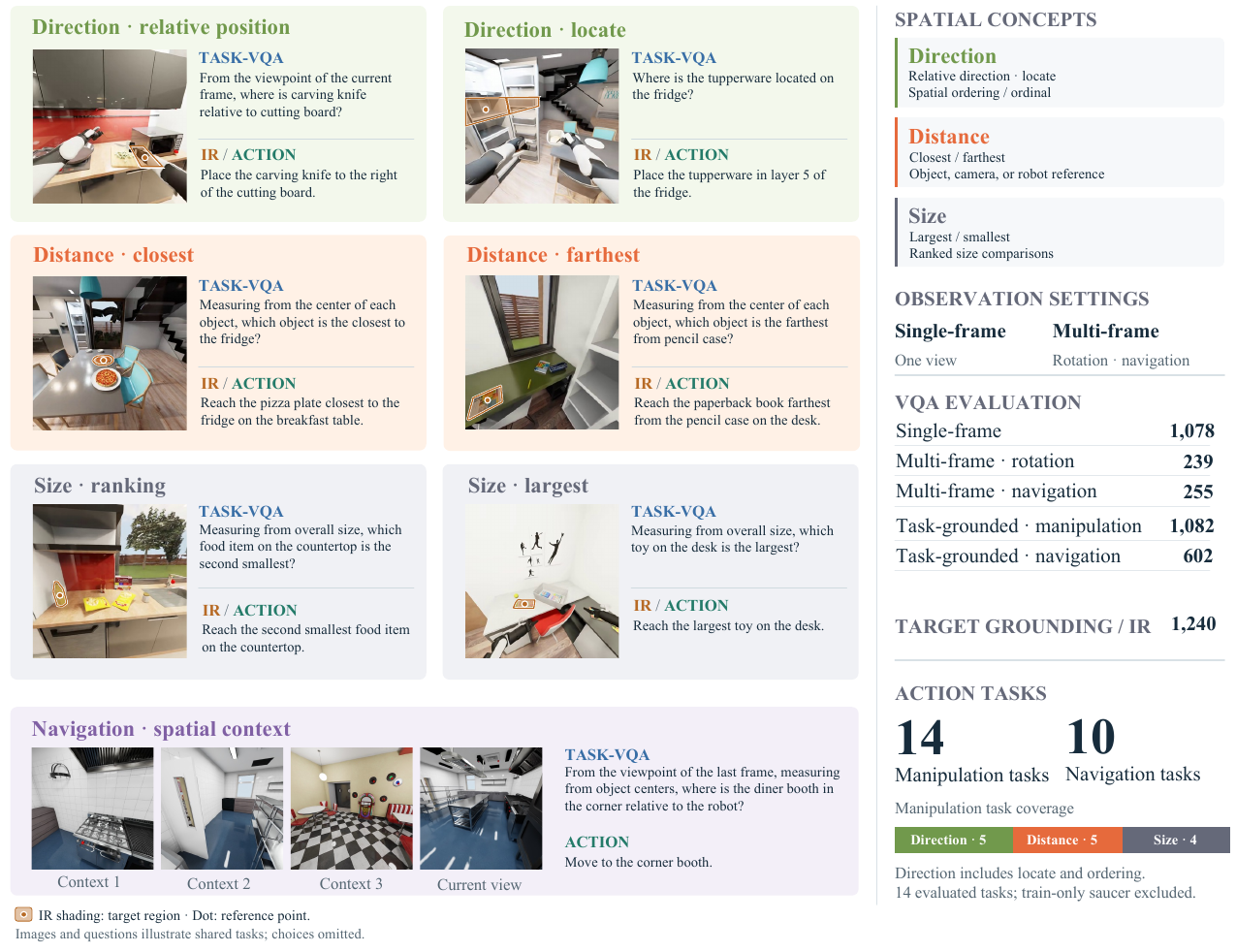}
    \vspace{-1em}
    \caption{\textbf{\benchname Dataset Showcase. } Left: Paired task VQA, grounding, and action examples cover direction, distance, and size; navigation shows three context views and the current view. Right: Spatial taxonomy, VQA counts, grounding/IR representations, and task coverage.}
    \vspace{-1em}
    \label{fig:dataset_showcase}
\end{figure*}

%% file: fig_text/fig_data_curation.tex
\begin{figure*}[ht]
    \centering
    \includegraphics[width=0.98\textwidth]{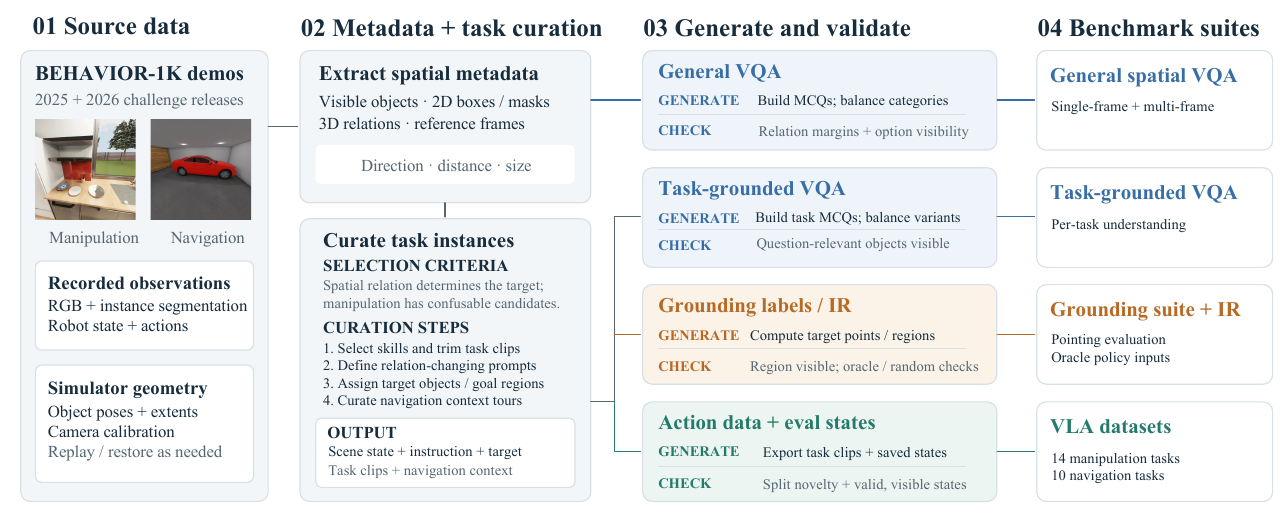}
    \vspace{-0.1cm}
    \caption{\textbf{Data Construction Pipeline.} Demonstration observations and simulator geometry support general VQA and curated spatial tasks. Task instances produce task-grounded VQA, grounding labels, and VLA datasets. Selection criteria guide curation; each suite validates its samples.}
    \vspace{-0.3cm}
    \label{fig:curation_pipeline}
\end{figure*}

%% file: text/sec4_vla_model.tex
\section{VLA Model}

\subsection{Model Architecture}
\label{sec:vla-model}

\vlaname builds on the $\pi_{0.5}$ architecture~\citep{intelligence2025pi05visionlanguageactionmodelopenworld},
replacing its VLM backbone with
Qwen3.5-4B~\citep{qwen3.5-4b}.
The VLM encodes visual observations and the language
instruction, with discretized robot proprioception
included in the text prefix. A flow-matching action
expert attends to the resulting prefix features
and predicts 32-step chunks of 23-dimensional
actions controlling the torso, dual arms, grippers,
and mobile base. Unless otherwise stated, the VLM
is frozen and the action expert is trained from
scratch.

For manipulation, policies are trained on
\datasetname's manipulation demonstrations and
receive head- and two wrist-camera views.
IR-conditioned variants additionally receive
oracle grounding images through the same visual
encoding interface. For navigation, policies
receive the current head-camera view and a
variable-length sequence of curated scene-context
frames, together with the instruction and robot
state.

\input{fig_text/fig_model_arch}






\subsection{Policy Variants}
\label{sec:policy-var}

We compare six manipulation policies, all conditioned
on camera observations, the instruction, and robot
state:
(1)~\textbf{Baseline} (frozen VLM);
(2)~\textbf{+LoRA-VLM}, which adapts the vision
and language components through LoRA;
(3)~\textbf{+DROID pretrain init}, which initializes
from DROID pretraining before task finetuning;
(4)~\textbf{+dot IR}, which adds a target-marked
head-camera RGB image;
(5)~\textbf{+dot IR + first-frame anchor}, which
also supplies a fixed, target-marked initial
head view; and
(6)~\textbf{+attn-map IR}, which uses a standalone
Gaussian heatmap in place of the dot-marked image. 

These variants test backbone adaptation, robot-data
transfer, and alternative grounding interfaces.
The dot and attention-map comparison changes both
spatial spread and image background. For navigation, we train a single frozen-VLM policy jointly on the 10 tasks. Appendices~\ref{app:policy-implementation}
and~\ref{app:policy-training} detail adaptation, initialization, and training.
Appendix~\ref{app:oracle-ir} specifies IR construction.

\subsection{Policy Evaluation Protocol}
\label{sec:eval-protocol}

\paragraph{Held-out \& Prompt-probe evaluation.}
The held-out set ($n{=}469$) uses scene configurations
from unseen demonstrations, paired with unseen
instruction variants that reverse the spatial
relation used during training. This setting tests
generalization under changes to both scene
configuration and instruction.
The prompt-probe set ($n{=}700$) retains scene
configurations used during training but changes
the instruction to specify a different valid
target. This setting tests whether policies follow the instruction when it conflicts with the demonstrated target.\looseness=-1

\paragraph{Manipulation evaluation.}
All manipulation policies are evaluated on the same instances.
Episodes use per-task step budgets calibrated
to the corresponding demonstration horizons
and are scored using the metrics defined in
\S\ref{sec:metrics}. For IR-conditioned policies,
we provide ground-truth grounding references
computed from simulator state. This oracle
setting evaluates action performance given
correct grounding, independently of errors
in VLM grounding predictions.

\paragraph{Navigation evaluation.}
We evaluate the jointly trained 10-task navigation
policy on spawn-validated held-out instances.
Appendix~\ref{app:nav-eval} describes instance
construction and visualizes initial poses and
scene layouts.
Training settings and evaluated checkpoints
are provided in Appendix~\ref{app:policy-training}.


\subsection{Code-as-policy agents}
\label{cap-agents}
Inspired by \cite{fu2026capxframeworkbenchmarkingimproving}, we additionally evaluate code-as-policy agents, in which a writer language model generates a Python program that controls the robot through a perception and control API including base and end-effector motion primitives, gripper control, RGB-D observations, and shared perception tools (an open-vocabulary pointing model and a VLM). Programs never receive privileged object poses, identities, or segmentation. For each task, the writer iteratively develops a program over 10 training episodes, disjoint from the test set, using execution traces, final observations, and simulator-derived outcome feedback. A validation-and-fallback procedure selects a program, which is then frozen and executed without writer revisions on all test episodes. Programs are evaluated on the same held-out episodes and with the same success metrics as the VLA policies. Appendix~\ref{app:cap} describes the program-development details, execution settings, and per-task results.

%% file: fig_text/fig_model_arch.tex
\begin{figure*}[ht]
    \centering
    \includegraphics[width=0.9\textwidth]{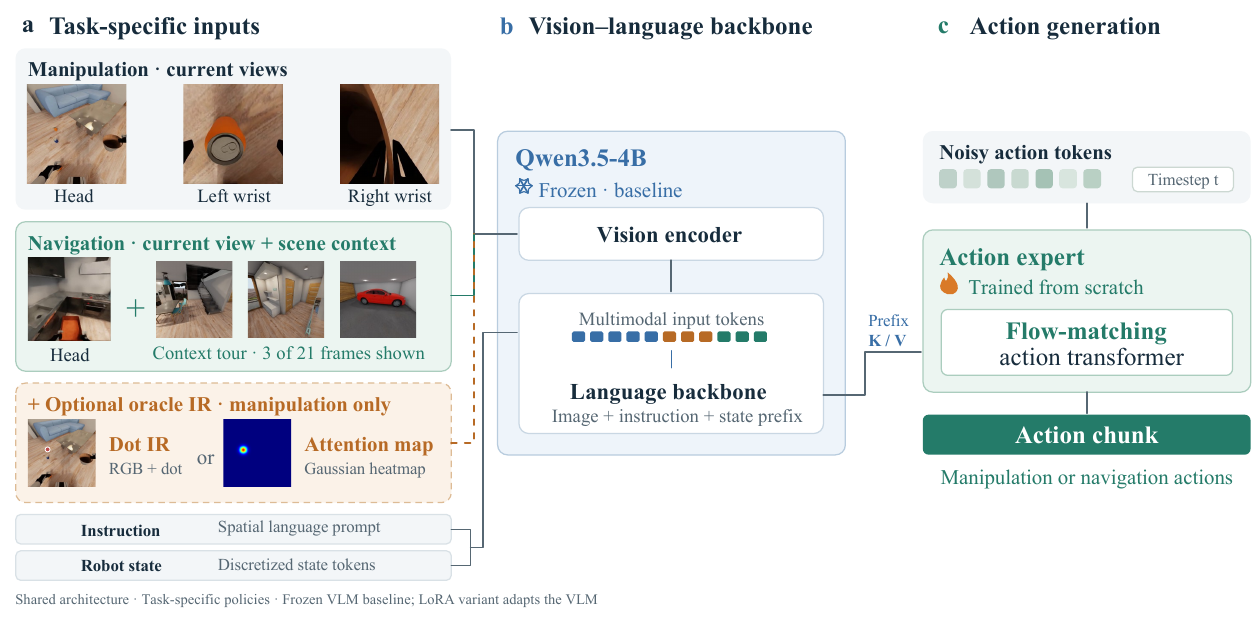}
    \vspace{-0.1cm}
    \caption{\textbf{\vlaname Architecture.} Task specific views share a Qwen3.5-4B backbone and flow matching action expert. Only manipulation uses optional oracle grounding.}
    \vspace{-0.3cm}
    \label{fig:model_arch}
\end{figure*}

%% file: text/sec6_experiments_results.tex
\section{Experiments and Results}
\label{sec:experiments}

\input{text/sec5_metrics}

\subsection{VLM Evaluation on \vqaname and \IRname}
\label{sec:vlm-vqa-ir-eval}

\paragraph{How well do VLMs answer spatial questions?}
Table~\ref{tab:vlm-vqa} evaluates 14 VLMs, comprising five proprietary -- GPT-6~Astra~\citep{openai2026gpt6astra}, Gemini~3~\citep{google2025gemini3pro}, Fable5.1~\citep{anthropic2026fable51card}, GPT-5~\citep{singh2026openaigpt5card}, and
Gemini~2.5~\citep{comanici2025gemini25pushingfrontier}-- and nine open-weight models spanning Qwen~\citep{qwen3.5}, Molmo~\citep{deitke2024molmopixmoopenweights}, Gemma~\citep{gemmateam2025gemma3}, and DeepSeek~\citep{wu2024deepseekvl2mixtureofexpertsvisionlanguagemodels}
families. VQA evaluation prompts, including spatial conventions
and multi-frame context instructions, are provided in
Appendix~\ref{app:vqa-prompts}.
The highest single-frame accuracy is 82.1\%, compared
with 51.8\% for the strongest open-weight model.
Multi-frame performance varies substantially across models, with
several open-weight models at or below the random-guess baseline on direction questions. Spatial VQA thus remains challenging, especially across multiple views. Extracted VLA backbones also score
poorly (Appendix~\ref{app:backbone-vqa-diagnostic}).


\paragraph{Does general VQA performance transfer to task-grounded questions?}
We evaluate the best open-source models from the general VQA comparison on task-grounded questions from the manipulation suite (Table~\ref{tab:task-perception}); only open-source models are evaluated, as only these can serve as our VLA backbones. Qwen3.5-9B
achieves the highest accuracy at 57.2\%, followed by Molmo2-4B at 53.0\% and Molmo2-8B at 50.0\%. Even Qwen3.5-9B attains 56.3\% on direction and 61.4\% on distance, but only 47.3\% on locate sub-types of direction questions. Strong general VQA thus does not guarantee task-grounded perception, motivating task-grounded evaluation of candidate backbones.
\paragraph{How accurately can VLMs ground the instructed target?}
The IR evaluation asks models to predict a target point directly from the instruction and head-camera view (Table~\ref{tab:task-perception}), using the prompts provided in Appendix~\ref{app:ir-prompts}. Qwen3.5-9B achieves the highest strict accuracy at 44.8\%, followed by Molmo2-8B at 44.6\% and Qwen3.5-4B at 38.6\%. Under tolerant scoring, these accuracies increase to 49.4\%, 52.2\%, and 46.6\%, respectively. Molmo2-4B and Molmo2-8B reverse order between VQA and pointing, so answer selection and localization probe different aspects of perception. We relate these task scores to policy behavior in
\S\ref{sec:perception-success}; qualitative examples appear in Appendix~\ref{app:perception-examples}.

\relax

\begin{table}[!h]\centering\begin{minipage}[t]{0.59\textwidth}\vspace{0pt}\input{fig_text/tab_task_vqa_ir_acc}\end{minipage}\hfill\begin{minipage}[t]{0.39\textwidth}\vspace{0pt}\input{fig_text/tab_spearman_corr}\end{minipage}\end{table}

\subsection{VLA evaluation on robot tasks}
\label{sec:exp-findings}

We evaluate the policy variants of \S\ref{sec:policy-var} on the manipulation and navigation tasks of \datasetname{} (Tables~\ref{tab:main-results} and~\ref{tab_nav-results}). Checkpoints are selected around similar training steps for fair comparisons.

The policy predicts 32-step action chunks using
10 flow-matching denoising steps per inference. Execution uses asynchronous receding-horizon control with a 10-step replanning threshold measured from the committed plan's input-observation timestamp. Actions are temporally ensembled over up to three committed plans, with gripper commands taken from the newest plan. Details are in Appendix~\ref{app:execution-settings}.


\subsubsection{VLA performance on manipulation tasks}
Results are task-dependent, with particularly limited instruction following when the requested target changes in training layouts. VLM adaptation and robot-data pretraining offer limited gains, while oracle target grounding yields larger improvements.

\begin{wraptable}{R}{0.55\textwidth}\input{fig_text/tab_vla_main_res}\end{wraptable}\paragraph{How well do policies act toward spatially specified targets?}
Pooled across manipulation tasks, on held-out instances the baseline achieves
41.6\% soft and 27.9\% strict success, with 38.2\%
wrong targets.

The baseline's per-task results
(Appendix~\ref{app:failure-analysis}) reveal distinct bottlenecks.
For \texttt{place\_carving\_knife}, 70.8\% soft success versus
29.2\% strict success suggests execution difficulty
despite frequent target-directed progress. Bowl placement achieves
only 15.0\% on both measures, with a 77.5\% wrong-target rate,
indicating that target selection itself is difficult.
Both target selection and execution thus limit completion.\looseness=-1

\paragraph{Do policies follow changed instructions in training layouts?}
The baseline often fails to redirect its behavior toward a newly
requested target, even in training layouts. On prompt-probe episodes,
it achieves only 16.1\% soft and 10.0\% strict success, while
70.0\% of episodes commit to a non-instructed target. The policy relies on scene--action associations rather than the instruction. In the failed probe example in
Figure~\ref{fig:vla-samples}, the instruction requests the largest
food item, but the policy reaches the chip bag associated with the
original training instruction rather than the indicated oatmeal box. Held-out target-directed behavior thus does not imply reliable instruction following.

\paragraph{Does VLM adaptation or robot-data pretraining improve instruction following?}
LoRA adaptation increases held-out soft success from 41.6\% to
43.5\%, but yields only a small change on prompt-probe episodes
(16.1\% to 17.9\%), with 68.1\% wrong-target commitments remaining.
DROID initialization achieves 40.1\% held-out and 16.6\% prompt-probe
soft success. Neither recipe improves strict probe success over
the baseline (9.6\% and 8.9\%, respectively, versus 10.0\%).

\paragraph{Does explicit target grounding improve policy performance?}
Oracle IR produces substantially larger gains: dot-IR conditioning
raises soft success from 41.6\% to 60.3\% on held-out set and
from 16.1\% to 38.0\% on the prompt probe. Across the three IR variants,
probe wrong-target commitments decrease from 70.0\% to 42.4--46.0\%.
Strict success also improves, reaching 44.8\% held-out and 24.9\% on the probe for the strongest variant with dot IR. Explicit grounding thus redirects policy behavior and improves completion.

\subsubsection{Code-as-policy agents on manipulation tasks}
\label{sec:cap-mani}
\begin{wraptable}{R}{0.4\textwidth}\input{fig_text/fig_cap_results}\end{wraptable} \paragraph{Do code-as-policy agents act toward spatially specified targets?} With the same perception tools and motion API, the proprietary writers achieve approximately 50–66\% soft success, compared with 37–39\% for the two open-weight writers. Thus, target-consistent behavior varies substantially with the model composing perception and control into executable programs. However, soft success does not imply reliable completion: Qwen3.5-4B and Qwen3.5-9B achieve only 5.1\% and 5.3\% strict success, respectively, whereas Gemini 3, GPT-6, and Fable 5.1 achieve 33.3\%, 47.3\%, and 51.4\%. See Table~\ref{tab:e2-per-task} for per-task results.
\paragraph{What failures remain?} The open-weight writers frequently produce partial target-directed behavior without completing an interaction, with 83–87\% of episodes ending without a decision. The proprietary writers complete interactions more frequently, but wrong-target commitments remain substantial. 

\subsubsection{Perception--Action correlation for manipulation}
\label{sec:perception-success}
\paragraph{Do perception scores track spatially directed behavior?}
We compare Qwen3.5-4B's task-level VQA and tolerant IR accuracies with
each policy's soft held-out success across 14 manipulation tasks,
using Spearman's rank correlation with average ranks for ties
(Table~\ref{tab:perception_action_correlation}). For two of the
strongest-performing variants, dot IR and attention-map IR, grounding
accuracy correlates more strongly with success than VQA accuracy ($\rho=0.701$ versus $0.560$, and $0.753$ versus $0.608$, respectively). This suggests that the metric that tracks the policy is the one the VLM has to generate, not the one it has to answer. Correlation plots are provided in Appendix~\ref{app:correlation-plots}.

\subsubsection{Ablation Studies for manipulation tasks}
\label{sec:exp-ablations}
We examine how target cues and demonstration design affect manipulation performance.

\paragraph{Does the IR rendering affect target-consistent behavior?}
We compare two ways of presenting the same oracle target center: a dot on the head-camera RGB image and a standalone Gaussian attention map. Both yield similar soft success on held-out and probe sets, but dot IR achieves higher strict success, possibly because it preserves the surrounding RGB context.

With VLM-predicted instead of oracle grounding, the dot-IR
policy retains part of this benefit: Qwen3.5-4B predictions yield 48.8\% held-out soft success, compared with 41.6\% for the baseline. Performance remains below oracle grounding (60.3\%). Appendix~\ref{app:predicted-ir} provides the rollout protocol and full results, including held-out comparisons across three grounding VLMs.

\paragraph{Does a first-frame reference improve performance?}
The target can leave the head-camera view as the robot moves toward it,
motivating a fixed reference that retains the initial scene and target
mark. We add the target-marked first frame as a fifth image input to the dot IR policy, held constant throughout the rollout. Both soft and strict success decrease slightly, so a persistent initial view adds no benefit over the current-view cue.

\paragraph{Does changing the demonstration distribution affect target selection?}
To examine the influence of demonstration geometry, we study the \texttt{reach\_paperback\_book} task in which the referred book is associated with a fixed side in the training data. We prefix each demonstration with motion from the two candidate books' perpendicular bisector to the original start, changing the approach geometry while retaining the original interaction. The main 14-task evaluation uses the augmented paperback demonstrations. On the shared prompt-probe set, this augmentation reduces wrong-target commitments from 79.7\% to 43.5\% and raises soft probe success from 11.6\% to 36.2\%, while strict success remains 10.1\%. This intervention changes the failure distribution without improving completion. Details are in Appendix~\ref{app:paperback-bias}.


\paragraph{How much does performance depend on demonstration volume?}
We retrain our best-performing variant, the dot-IR policy, on
nested 30\% and 50\% subsets of the demonstrations, preserving
the task mixture and training for the same number of updates.
Increasing the data from 30\% to 100\% raises held-out soft/strict success from 56.1\%/34.5\% to 60.3\%/44.8\%, and prompt-probe soft/strict success from 32.6\%/19.9\% to 38.0\%/24.9\%.
The full-data policy performs best overall, although gains are
not monotonic: the 50\% subset yields similar or lower success
than the 30\% subset. Even with 30\% of the demonstrations,
dot IR substantially outperforms the full-data frozen-VLM
baseline on both evaluation sets.
See Appendix~\ref{app:datascaling} for details.

\subsection{VLA performance on navigation tasks}
\label{sec:exp-nav}

\paragraph{How does navigation success vary across tasks?}
Success ranges from 22.5\% for the garage to 92.5\% for the bookcase (Table~\ref{tab_nav-results}). The bookcase and bathroom tasks achieve
high success on short-range navigations, while the garage task remains
challenging. The diner scene and two-floor house also exhibit substantial variation across
destinations, with success from 25--35\% to 75--85\%. Scene identity alone
therefore does not account for the task ordering. Reaching a nearby
object or region and navigating across a larger environment impose different demands on context use and motion. Route length, obstacles,
and doorway crossings are plausible contributors to these differences. See Appendix~\ref{app:qualitative-policy}.

\paragraph{Does navigation VQA accuracy track policy success?}
We evaluate Qwen3.5-4B on multi-frame direction-question for navigation, accuracy ranges from 16.7\% to 40.0\%. Across the ten tasks, the weak correlation with navigation success indicates that VQA accuracy alone does not explain variation in task success. Navigation-VQA examples are provided in
Figure~\ref{fig:nav_vqa}.
The complete results for all ten navigation tasks are reported in Table~\ref{tab_nav-results} (Appendix~\ref{app:navigation-results}). The table pairs VQA accuracy and policy success by task and scene to show variation across environments.

%% file: text/sec5_metrics.tex
\subsection{Metrics}
\label{sec:metrics}
\paragraph{VQA and IR Accuracy.}
VQA accuracy measures the fraction of questions answered correctly under the four-option evaluation interface; unparseable responses count as incorrect. For IR evaluation, we ask model to predict  a target point, measuring strict IR accuracy uses task-specific target-region criteria and tolerant IR accuracy that additionally allow small localization errors for reach tasks. We report task-grounded manipulation VQA and IR accuracy as averages of per-task accuracies. Detailed scoring rules, tolerances, and aggregation conventions are provided in Appendix~\ref{app:metrics}.

\begin{wraptable}{r}{0.55\textwidth}\vspace{-10pt}\input{fig_text/tab_general_vqa_acc}\vspace{-10pt}\end{wraptable}
\paragraph{Robot Task Success.}
For manipulation, we primarily use \emph{soft success} as a proxy for target-selection intention. It credits completed correct interactions as well as non-committed episodes that exhibit correct intention for target approaching under task-specific rules and timeout budgets. We also report \emph{strict success} for task completion rate, and the \emph{wrong-target rate} measuring wrong commitments to non-instructed candidates or locations. All three rates report percentage over all evaluated episodes. Task-family rules and exceptions are detailed in Table~\ref{tab:success-criteria}. Navigation uses the recorded binary goal result: the robot must
satisfy the instructed region-entry or object-proximity criterion
within the episode timeout budget. We report per-task and pooled success, without the strict/soft distinction.

%% file: fig_text/tab_general_vqa_acc.tex
\vspace{-1em}
\caption{\textbf{VLM accuracy on \vqaname.} Single-frame Direction
includes its locate sub-type; All is pooled over all questions.}
\label{tab:vlm-vqa}
\centering\setlength{\tabcolsep}{2pt}
\footnotesize
\resizebox{\linewidth}{!}{\begin{tabular}{@{}lcccccc@{}}
\multicolumn{1}{c}{\bf Model}
& \multicolumn{4}{c}{\bf Single-frame}
& \multicolumn{2}{c}{\bf Multi-frame} \\
\multicolumn{1}{c}{}
& All & Dir. & Dist. & Size & All & Dir.
\\ \hline
\multicolumn{7}{l}{\it Proprietary} \\[2pt]
GPT-6 Astra         & {\bf 82.1} & {\bf 85.8} & {\bf 77.3} & {\bf 77.7} & {\bf 74.4} & {\bf 74.6} \\
Gemini 3            & 77.8 & 80.5 & 75.2 & 72.7 & 71.1 & 69.6 \\
Fable5.1            & 70.5 & 69.3 & 70.2 & 76.3 & 68.4 & 66.8 \\
GPT-5               & 64.7 & 61.7 & 68.1 & 69.1 & 63.2 & 59.6 \\
Gemini 2.5          & 59.9 & 55.0 & 65.2 & 68.3 & 60.1 & 57.4 \\[4pt]
\multicolumn{7}{l}{\it Open-weight} \\[2pt]
Molmo2-8B           & 51.8 & 45.0 & 57.5 & 66.9 & 36.8 & 30.7 \\
Qwen3.5-9B       & 50.6 & 42.0 & 58.1 & 69.1 & 54.5 & 51.4 \\
Qwen3.5-4B       & 46.8 & 44.8 & 48.1 & 51.8 & 46.6 & 44.1 \\
Molmo2-4B           & 45.5 & 34.2 & 58.7 & 62.6 & 32.4 & 23.7 \\
PaliGemma           & 36.2 & 32.5 & 40.7 & 41.0 & 37.9 & 34.4 \\
PaliGemma\,2        & 34.8 & 29.8 & 39.2 & 45.3 & 27.3 & 23.4 \\
Gemma\,3-4B         & 34.4 & 35.0 & 33.6 & 33.8 & 29.2 & 21.3 \\
Qwen2.5-VL          & 33.4 & 31.2 & 37.8 & 32.4 & 28.9 & 22.9 \\
DeepSeek-VL2-Tiny   & 31.4 & 30.5 & 34.8 & 26.6 & 25.0 & 17.6 \\[4pt]
\multicolumn{7}{l}{\it VLM backbones extracted from released VLA models} \\[2pt]
OpenHelix           &  9.6 &  5.8 & 14.2 & 15.1 & ---  & --- \\
$\pi_{0.5}$         & 0.9 & 0.8 & 0.3 & 2.9 & ---  & --- \\
\hline
Chance              & 25.0 & 25.0 & 25.0 & 25.0 & 25.0 & 25.0 \\
\end{tabular}}
\centering\setlength{\tabcolsep}{2pt}
\vspace{-1em}

%% file: fig_text/tab_task_vqa_ir_acc.tex
\caption{\textbf{Task-grounded perception performance for manipulation.}
VQA and point-grounding accuracy (\%) for the same VLMs.
VQA reports overall and category-level accuracy; IR reports
strict and tolerant accuracy. Higher is better in all columns; bold indicates the highest accuracy.}
\label{tab:task-perception}
\centering
\footnotesize
\setlength{\tabcolsep}{2.5pt}
\begin{tabular}{lcccc@{\hspace{7pt}}cc}
\multicolumn{1}{c}{\bf Model}
& \multicolumn{4}{c}{\bf Task-grounded VQA}
& \multicolumn{2}{c}{\bf IR} \\
& All & Dir. & Dist. & Size & Strict & Tolerant
\\ \hline
Qwen3.5-9B & {\bf 57.2} & {\bf 56.3} & {\bf 61.4} & {\bf 52.5} & {\bf 44.8} & 49.4 \\
Molmo2-4B  & 53.0 & 54.1 & 56.4 & 46.8 & 26.8 & 39.5 \\
Molmo2-8B  & 50.0 & 49.9 & 53.4 & 45.3 & 44.6 & {\bf 52.2} \\
Qwen3.5-4B & 49.4 & 50.9 & 48.6 & 48.6 & 38.6 & 46.6 \\
Gemma\,3-4B & 33.5 & 23.0 & 42.4 & 36.0 & 34.1 & 44.1 \\
\vspace{-2em}
\end{tabular}

%% file: fig_text/tab_spearman_corr.tex
\caption{\textbf{Task-level perception--action correlation.} Spearman correlations across 14 manipulation tasks between VQA or tolerant IR accuracy and policy success. Ties receive average ranks.}
\label{tab:perception_action_correlation}
\centering
\footnotesize
\setlength{\tabcolsep}{2.5pt}
\begin{tabular}{lcc}
\multicolumn{1}{c}{\textbf{Policy}} & \textbf{VQA} ($\rho$) & \textbf{IR} ($\rho$) \\
\hline
Baseline & 0.713 & 0.658 \\
LoRA VLM & 0.616 & 0.581 \\
IR & 0.560 & 0.701 \\
IR + first-frame IR & 0.612 & 0.618 \\
Attention map & 0.608 & 0.753 \\
\vspace{-3em}
\end{tabular}

%% file: fig_text/tab_vla_main_res.tex

\caption{\textbf{Success rates of all policy variants on the two evaluation sets.} +Soft = soft success; S = strict success; W = fraction of episodes committing to a
non-instructed object. 
}
\label{tab:main-results}
\centering\footnotesize\setlength{\tabcolsep}{2pt}
\resizebox{\linewidth}{!}{\begin{tabular}{@{}lccccc@{}}
\multicolumn{1}{c}{\bf Policy}
& \multicolumn{2}{c}{\bf Held-out}
& \multicolumn{3}{c}{\bf Prompt-probe} \\
\multicolumn{1}{c}{}
& +Soft$\uparrow$ & S$\uparrow$
& +Soft$\uparrow$ & S$\uparrow$ & W$\downarrow$
\\ \hline
$\pi_{0.5}$-Qwen (frozen VLM)           & 41.6 & 27.9 & 16.1 & 10.0 & 70.0 \\
\quad + LoRA-VLM                        & 43.5 & 30.9 & 17.9 &  9.6 & 68.1 \\
\quad + DROID pretrain init             & 40.1 & 24.7 & 16.6 &  8.9 & 66.0 \\[4pt]
\multicolumn{6}{l}{\it IR-conditioned} \\[2pt]
\quad + dot IR                          & {\bf 60.3} & {\bf 44.8} & {\bf 38.0} & {\bf 24.9} & 43.9 \\
\quad + dot IR + first-frame anchor     & 59.1 & 41.8 & 36.3 & 23.7 & 42.4 \\
\quad + attn-map IR                     & 59.3 & 39.0 & 35.0 & 17.0 & 46.0 \\ [4pt]
\multicolumn{6}{l}{\it Data scaling (\S\ref{app:datascaling})} \\[2pt]
\quad + dot IR @ 50\% data               & 53.7 & 33.7 & 30.6 & 19.4 & 45.7 \\
\quad + dot IR @ 30\% data               & 56.1 & 34.5 & 32.6 & 19.9 & 46.6 \\
\end{tabular}}
\vspace{-2em}
\centering\footnotesize\setlength{\tabcolsep}{2pt}

%% file: fig_text/fig_cap_results.tex
\vspace{-2em}
\caption{\textbf{Code-as-policy agent performance.} +Soft = soft success; S = strict success; W = fraction of episodes committing to a
non-instructed object. }
\label{tab:e2-results}
\centering
\footnotesize
\setlength{\tabcolsep}{3pt}
\begin{tabular}{@{}lccc@{}}
\multicolumn{1}{c}{\bf Writer}
& +Soft$\uparrow$ & S$\uparrow$ & W$\downarrow$
\\ \hline
\multicolumn{4}{l}{\it Proprietary} \\[2pt]
Fable 5.1     & {\bf 66.1} & {\bf 51.4} & 23.5 \\
GPT-6         & 61.2 & 47.3 & 23.0 \\
Gemini 3 & 50.3 & 33.3 & 28.8 \\[4pt]
\multicolumn{4}{l}{\it Open-weight} \\[2pt]
Qwen3.5-9B & 39.0 & 5.3 & 8.7 \\
Qwen3.5-4B & 37.1 & 5.1 & 7.2 \\
\end{tabular}
\vspace{-2em}

%% file: text/sec7_limitations.tex
\section{Limitations and Conclusion}
\subsection{Limitations}
Our evaluation is conducted entirely in simulation; transfer to physical
robots remains to be established. In the tupperware task, correlations
between the acting hand and the target layer can confound spatial
instruction following with hand-specific manipulation behavior.
Success rates also depend on the episode timeout budget, so cross-run
comparisons should account for differences in the available execution
time. Finally, the IR-conditioned policies require simulator-derived
IR target centers during training; applying this training approach
to datasets without such annotations would require an alternative
source of grounding supervision.\looseness=-1

\subsection{Conclusion}
We introduced \benchname to jointly evaluate spatial question
answering, target grounding, and robot action in household tasks. The evaluated policies often fail to act on the instructed target, particularly when prompts change in familiar layouts. Explicit IR target cues improve target-directed behavior, with gains persisting under VLM-predicted grounding, although reliable completion remains challenging. Perception scores offer policy-dependent diagnostics but do not consistently explain action success. By distinguishing
target selection, partial progress, and task completion, \benchname provides a framework to measure when spatial understanding supports successful action and when it fails.\looseness=-1

%% file: supp.tex
\clearpage
\renewcommand\thefigure{S\arabic{figure}}
\setcounter{figure}{0}
\renewcommand\thetable{S\arabic{table}}
\setcounter{table}{0}
\renewcommand\theequation{S\arabic{equation}}
\setcounter{equation}{0}
\pagenumbering{arabic}
\renewcommand*{\thepage}{S\arabic{page}}
\setcounter{footnote}{0}
\setcounter{page}{1}
\renewcommand{\theHfigure}{supp.\arabic{figure}}
\renewcommand{\theHtable}{supp.\arabic{table}}
\renewcommand{\theHequation}{supp.\arabic{equation}}

\appendix

\input{supp_text/appx_data_curation_details}
\input{supp_text/appx_task_suite_overview}

\input{supp_text/appx_model_details}

\input{supp_text/appx_metrics}

\input{supp_text/appx_vqa_eval}

\input{supp_text/appx_experiment}

%% file: supp_text/appx_data_curation_details.tex
\section{Additional Data Curation Details}
\label{app:data-curation}

This section supplements the pipeline in \S\ref{sec:data-curation}
with spatial concepts, question curation and frame selection, and oracle IR construction.

\subsection{Spatial Concepts and Observation Settings}
\label{app:spatial-taxonomy}

We detail the three spatial concepts covered by
\benchname and the observation settings used
for spatial VQA.

\paragraph{Direction.}
Direction concerns the relative position of an entity
with respect to a reference, including left/right,
front/back, and above/below relations. These relations
are expressed in object- or robot-relative
reference frames. We also include two subtypes that
require finer-grained spatial identification:
\emph{locate} identifies the region of a container or
piece of furniture occupied by an object, such as a
shelf layer, fridge level, or bookcase grid cell;
\emph{ordinal} identifies an object by its rank in a
spatial ordering, such as ``the second leftmost tool.''
Ordinal questions appear only in the task-grounded
VQA subset, where they reflect the referring expressions
used in the robot tasks.

\paragraph{Distance.}
Distance concerns comparisons of physical separation
between entities. Questions and task instructions
identify the candidate closest to or farthest from
a reference object, the camera, or the robot arm.
Resolving these relations requires comparing distances
across candidates.

\paragraph{Size.}
Size concerns comparisons of physical size among
objects of the same category, such as identifying
the largest or smallest candidate. These comparisons
require accounting for perspective, since an object's
apparent size in the image depends on both its physical
size and its distance from the camera.

\paragraph{Observation setting.}
Single-frame questions evaluate spatial understanding
from one rendered view. Multi-frame questions require
integrating spatial information across viewpoints
and comprise two settings. In \emph{rotation}, the
robot rotates in place, and questions require relating
entities observed in different views within a common
egocentric frame. In \emph{navigation}, the robot moves
through a scene, and questions require tracking spatial
relations across observations while accounting for
the robot's displacement.

\input{fig_text/tab_vqa_templates}

\subsection{VQA Generation, Filtering, and Frame Selection}
\label{app:vqa-curation}

\vqaname comprises general and task-grounded spatial
questions, both generated through a metadata-to-template
pipeline. For each sampled frame, we extract 3D object
states, projected 2D bounding boxes, and robot reference
frames. A spatial calculator computes direction
relations in object- and robot-relative
frames; metric distances, including perpendicular
distances to support surfaces; size comparisons; and
containment relations for locate questions.
Per-category templates convert these relations into
four-option multiple-choice questions, with distractors
drawn from the corresponding objects or spatial
relations.

\paragraph{General VQA filtering and sampling.} Generation is followed by
three filtering and sampling steps. First,
relation-margin checks require at least a 15\%
relative separation for distance comparisons and
a 10\% margin for size comparisons. Second,
visibility checks filter queried and candidate
objects by their image pixel counts, with a stricter
threshold for the ground-truth object where applicable.
Third, a sublinear cap on questions per frame limits
the contribution of cluttered scenes. We then
subsample the resulting pool to control category
coverage, yielding 1,078 single-frame and 494
multi-frame questions. These checks reduce ambiguity
and sampling imbalance without guaranteeing that
every geometric distinction is visually recoverable. Table~\ref{tab:vqa-templates} summarizes the question
templates, answer spaces, and evaluation counts for
the general VQA subset.

\paragraph{Multi-frame inputs.}
General rotation questions supply four frames captured
as the robot rotates in place, jointly covering the
full $360^{\circ}$ surround. General navigation
questions use observations captured during movement
through a scene. Multi-frame direction questions
use four diagonal relations (front-left, front-right,
back-left, and back-right), whereas single-frame
direction questions use axis-aligned relations.

\paragraph{Task-grounded VQA.}
\taskvqaname uses task-conditioned templates to query
objects and spatial relations associated with
policy-evaluation instances. Frames are selected
from these instances and filtered for visibility
of the entities required by each question.
Question counts are balanced across instruction
variants within each task, yielding 1,082
manipulation questions and 602 navigation questions.
The perception and action suites are aligned at
the task level, while their valid evaluation
samples may differ because of suite-specific
visibility requirements.

\input{fig_text/fig_15mani_overview}
\input{fig_text/fig_10nav_overview}

\paragraph{Navigation context.}
Task-grounded navigation VQA receives curated
per-environment context frames together with the
current observation. Navigation policies receive
the same ordered context frames as variable-length
visual input (\S\ref{sec:vla-model}).
Question answers are computed from simulator state.
Appendix~\ref{app:task-suite} describes context
traversal construction and illustrates the resulting
frame sequences.

\subsection{Point-Grounding Annotation and Validation}
\label{app:ir-curation}



\paragraph{Evaluation frame selection.}
Evaluation frame counts are balanced across
instruction variants within each task. VQA and
point grounding apply distinct visibility criteria
to the same source frames because they require
different visual evidence. A VQA question such as
``Where is the knife relative to the frying pan?''
requires the knife and pan to be visible. A grounding
instruction such as ``Mark where the knife should go,
right of the pan'' additionally requires the
corresponding placement region to be visible.
If that region lies entirely outside the image,
there is no valid image-space answer.
Grounding evaluation therefore retains only frames
with a visible answer region, and its eligible
frame set may differ from that of VQA.

\paragraph{Evaluator validation.}
We use IR predictions and uniformly sampled image
points as complementary checks of the grounding
evaluator against its geometric acceptance regions.
Appendix~\ref{fig:ir_question_template} presents the grounding
query template.

\input{supp_text/tab_vla_tasks_inventory}

%% file: fig_text/tab_vqa_templates.tex
\begin{table}[htbp]
\caption{\textbf{\vqaname general subset templates.}
Question templates, answer spaces, and evaluation counts.}
\label{tab:vqa-templates}
\begin{center}
\scriptsize
\setlength{\tabcolsep}{3pt}%
\begin{tabular}{llL{5.8cm}L{2.9cm}c}
\multicolumn{1}{c}{\bf Setting}
& \multicolumn{1}{c}{\bf Category}
& \multicolumn{1}{c}{\bf Template}
& \multicolumn{1}{c}{\bf Answer space}
& \multicolumn{1}{c}{\bf $n$} \\
\hline
single-frame & direction
& ``From the viewpoint of \texttt{\{frame\_idx\}},
where is \texttt{\{objA\}} relative to
\texttt{\{objB / robot arm\}}?''
& left / right / front / back / above / below
& 300 \\
single-frame & locate
& ``From the viewpoint of \texttt{\{frame\_idx\}},
where is \texttt{\{objA\}} located on \texttt{\{objB\}}?''
& container region (layer/level/grid)
& 300 \\
single-frame & distance
& ``Measuring from the center of each object,
which object is the closest to
\texttt{\{objA / camera / robot arm\}}?''
& candidate objects
& 339 \\
single-frame & size
& ``Measuring from the length of the longest dimension
(length, width, or height), which object is the longest?''
& candidate objects
& 139 \\[3pt]
multi-frame & direction
& ``From the viewpoint of the last frame, measuring
from room centers, where is \texttt{\{objA\}}
relative to \texttt{\{objB\}}?''
& front-left / front-right / back-left / back-right
& 401 \\
multi-frame & distance
& ``Based on the given frames, measuring from room centers,
which room is the closest to the robot's position
at the last frame?''
& candidate rooms
& 77 \\
multi-frame & size
& ``Based on the given frames, which room has the largest area?''
& candidate rooms
& 16 \\
\end{tabular}
\smallskip
\begin{minipage}{\linewidth}
\scriptsize
\textit{Notes.}
Braces indicate placeholders; slash-separated entries indicate
alternative reference entities, not literal question wording.
Each question has four answer options;
$n$ denotes the number of evaluation questions.
Single-frame direction comprises 150 object-to-object and
150 robot-arm-relative questions; including the locate subtype,
it totals 600 questions.
Single-frame distance comprises 143 object-relative,
143 camera-relative, and 53 robot-arm-relative questions.
\end{minipage}
\end{center}
\end{table}

%% file: fig_text/fig_15mani_overview.tex
\begin{figure*}[ht]
    \centering
    \includegraphics[width=0.97\textwidth]{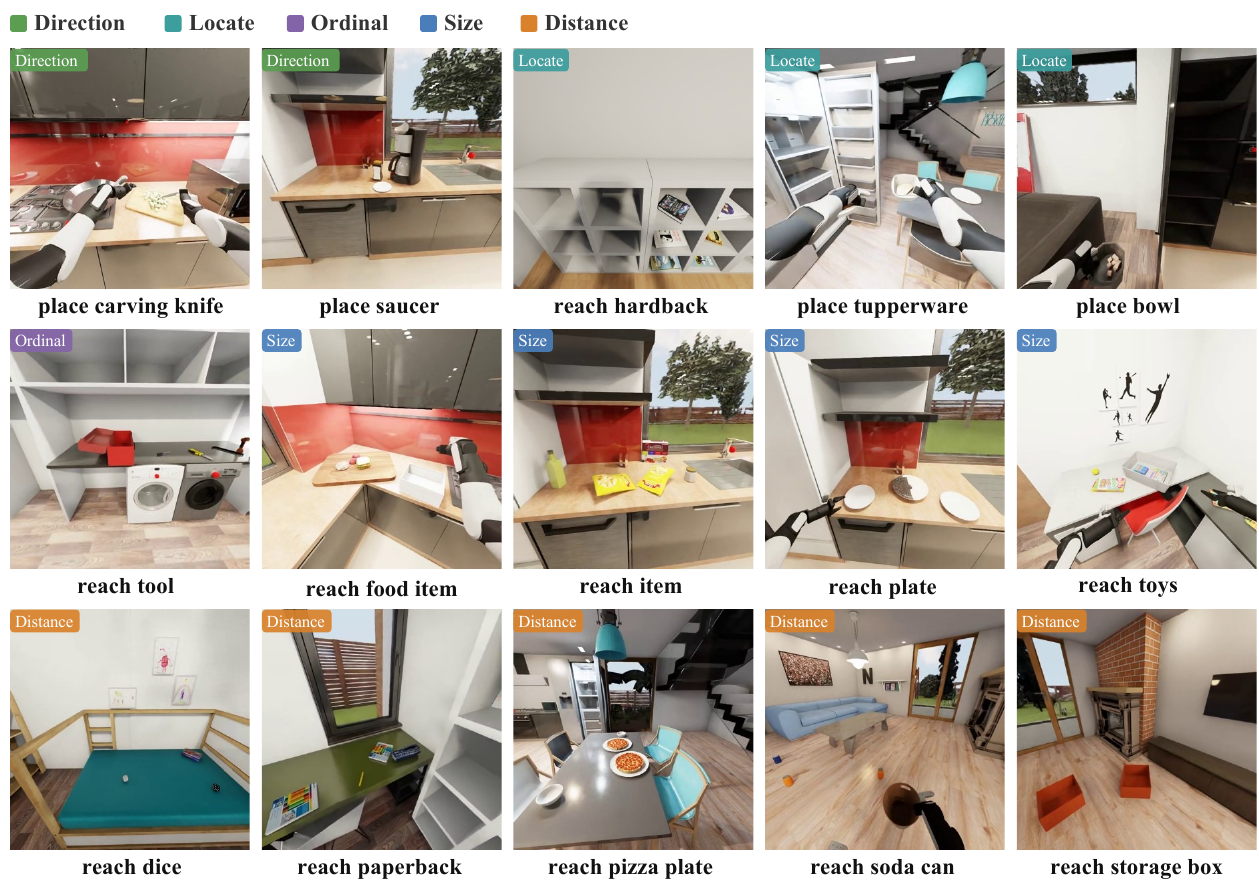}
    \vspace{-0.1cm}
    \caption{\textbf{Manipulation tasks of \benchname.}
    Representative head-camera views of the 14 tasks used
    for both training and evaluation, plus one training-only task.
    Color badges indicate the queried spatial concept.
    Table~\ref{tab:task-inventory} lists instructions and prompt variants.}
    \vspace{-0.3cm}
    \label{fig:15mani_overview}
\end{figure*}

%% file: fig_text/fig_10nav_overview.tex
\begin{figure*}[ht]
    \centering
    \includegraphics[width=0.97\textwidth]{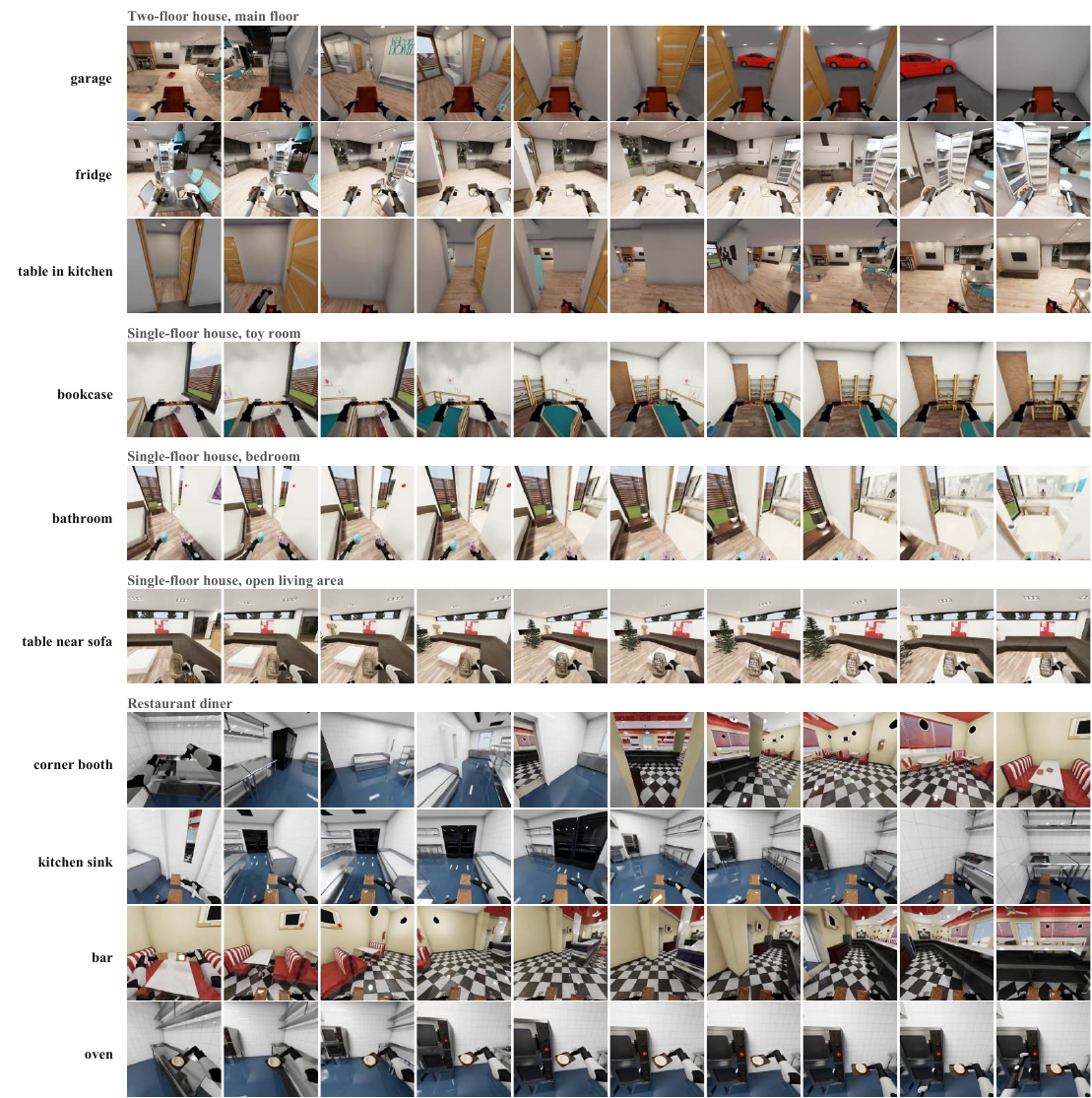}
    \vspace{-0.1cm}
    \caption{\textbf{The 10 navigation tasks of \benchname.}
    Each row shows ten evenly sampled frames from a source demonstration, from the starting view to the instructed target. Routes include in-room approaches, open-area traversals, and passage through doorways. Instructions follow the template ``Move to the \{target\}''
    (Table~\ref{tab:task-inventory}).}
    \vspace{-0.3cm}
    \label{fig:10nav_overview}
\end{figure*}

%% file: supp_text/tab_vla_tasks_inventory.tex
\begin{table}[htbp]
\caption{\textbf{Task inventory of \benchname.}
The benchmark comprises 14 manipulation tasks and 10 navigation
tasks; one additional manipulation task is used only for training.
Manipulation variants change the spatial relation or reference
object; $\times$ denotes all combinations.
Navigation evaluation varies validated starting poses rather than
instructions. $n$ denotes held-out evaluation instances.}
\label{tab:task-inventory}
\begin{center}
\scriptsize
\setlength{\tabcolsep}{3pt}%
\begin{tabular}{llL{4.6cm}L{3.2cm}c}
\multicolumn{1}{c}{\bf Task}
& \multicolumn{1}{c}{\bf Category}
& \multicolumn{1}{c}{\bf Example instruction}
& \multicolumn{1}{c}{\bf Instruction variants}
& \multicolumn{1}{c}{\bf $n$}
\\ \hline \\
\multicolumn{5}{l}{\it Manipulation} \\[2pt]
reach\_food\_item & size & ``Reach the largest food item on the chopping board.'' & \{largest, 2nd largest, smallest, 2nd smallest\} & 32 \\
reach\_item & size & ``Reach the smallest food item on the countertop.'' & \{largest, 2nd largest, smallest, 2nd smallest\} & 32 \\
reach\_plate & size & ``Reach the largest plate from the countertop.'' & \{largest, 2nd largest, smallest, 2nd smallest\} & 31 \\
reach\_toys & size & ``Reach the second largest toy on the desk.'' & \{largest, 2nd largest, 3rd largest, smallest, 2nd smallest\} & 40 \\
reach\_dice & distance & ``Reach the dice closest to the board game on the bed.'' & \{closest, farthest\} $\times$ \{board game, wall\} & 32 \\
reach\_paperback\_book & distance & ``Reach the paperback book farthest from the folder on the desk.'' & \{closest, farthest\} $\times$ \{folder, pencil case\} & 22 \\
reach\_pizza\_plate & distance & ``Reach the pizza plate closest to the fridge on the breakfast table.'' & \{closest, farthest\} $\times$ \{fridge, black chair, white chair\} & 40 \\
reach\_soda\_can & distance & ``Reach the can of soda farthest from the sofa on the floor.'' & \{closest, farthest\} $\times$ \{sofa, coffee table, bottom cabinet\} & 42 \\
reach\_storage\_box & distance & ``Reach the storage box closest to the coffee table on the floor.'' & \{closest, farthest\} $\times$ \{coffee table, bottom cabinet\} & 32 \\
reach\_tool & ordinal & ``Reach the tool on the second leftmost of the countertop.'' & \{leftmost, 2nd leftmost, rightmost, 2nd rightmost\} & 32 \\
reach\_hardback & locate & ``Reach the hardback book at the 2x2 grid of the bookcase.'' & grid cells \{1x1, 1x2, 2x1, 2x2, 3x1, 3x2\} & 48 \\
place\_carving\_knife & direction & ``Place the carving knife on the countertop and to the right of the cutting board.'' & \{left, right\} $\times$ \{cutting board, microwave, burner, frying pan\} & 24 \\
place\_saucer & direction & ``Place the saucer on the countertop and to the left of the electric kettle.'' & \{left, right\} $\times$ \{coffee maker, kettle, coffee bottle\} & train-only \\
place\_tupperware & locate & ``Place the tupperware in the layer 5 of the fridge.'' & layers \{4, 5, 6\} & 22 \\
place\_bowl & locate & ``Place the bowl in the layer 3 of the bottom cabinet.'' & layers \{1, 2, 3, 4, 5\} & 40 \\[4pt]
\multicolumn{5}{l}{\it Navigation (two-floor house)} \\[2pt]
move\_to\_garage & navigation & ``Move to the garage.'' & --- & 40 \\
move\_to\_fridge & navigation & ``Move to the fridge.'' & --- & 40 \\
move\_to\_table\_in\_kitchen & navigation & ``Move to the table in the kitchen.'' & --- & 40 \\[4pt]
\multicolumn{5}{l}{\it Navigation (single-floor house)} \\[2pt]
move\_to\_bookcase & navigation & ``Move to the bookcase.'' & --- & 40 \\
move\_to\_bathroom & navigation & ``Move to the bathroom.'' & --- & 40 \\
move\_to\_table\_near\_sofa & navigation & ``Move to the table near the sofa.'' & --- & 41 \\[4pt]
\multicolumn{5}{l}{\it Navigation (restaurant diner)} \\[2pt]
move\_to\_corner\_booth & navigation & ``Move to the corner booth.'' & --- & 40 \\
move\_to\_bar & navigation & ``Move to the bar.'' & --- & 40 \\
move\_to\_kitchen\_sink & navigation & ``Move to the kitchen sink.'' & --- & 40 \\
move\_to\_oven & navigation & ``Move to the oven.'' & --- & 40 \\
\end{tabular}
\vspace{-1em}
\end{center}
\end{table}

%% file: supp_text/appx_task_suite_overview.tex
\section{Task Suite Overview}
\label{app:task-suite}

The benchmark task suite comprises 14 manipulation
tasks and 10 navigation tasks. Figures~\ref{fig:15mani_overview} and~\ref{fig:10nav_overview} illustrate the manipulation task settings and navigation scene
layouts, respectively. Table~\ref{tab:task-inventory} lists all tasks and their instruction variants.

\paragraph{Manipulation task curation.}
We select manipulation subtasks in which a spatial
relation determines the intended target among
multiple plausible candidates. The tasks cover
direction, including locate and ordinal relations,
as well as distance and size. Demonstrations are
trimmed to the corresponding subtask windows.

\paragraph{Instruction variants.}
For each task, we curate instruction variants that
change the spatial relation or reference object
and assign the corresponding target object or
goal region to each variant. These variants support
the held-out and prompt-probe evaluations described
in \S\ref{sec:eval-protocol}.

\paragraph{Navigation tasks and context traversals.}
The navigation suite spans five scenes and requires
navigating to an instructed furniture target or region.
For each environment, we manually curate a traversal
covering relevant navigable regions and doorways
and collect a dense, ordered sequence of context
frames along the route. Figures~\ref{fig:nav-context} and~\ref{fig:nav-context-part2} show the complete context-frame sequences for all five scenes in input order. Question generation using these frames is described
in Appendix~\ref{app:vqa-curation}.

\input{fig_text/fig_nav_ctx}





%% file: fig_text/fig_nav_ctx.tex
\begin{figure*}[p]
    \centering
    \includegraphics[width=0.9\textwidth]{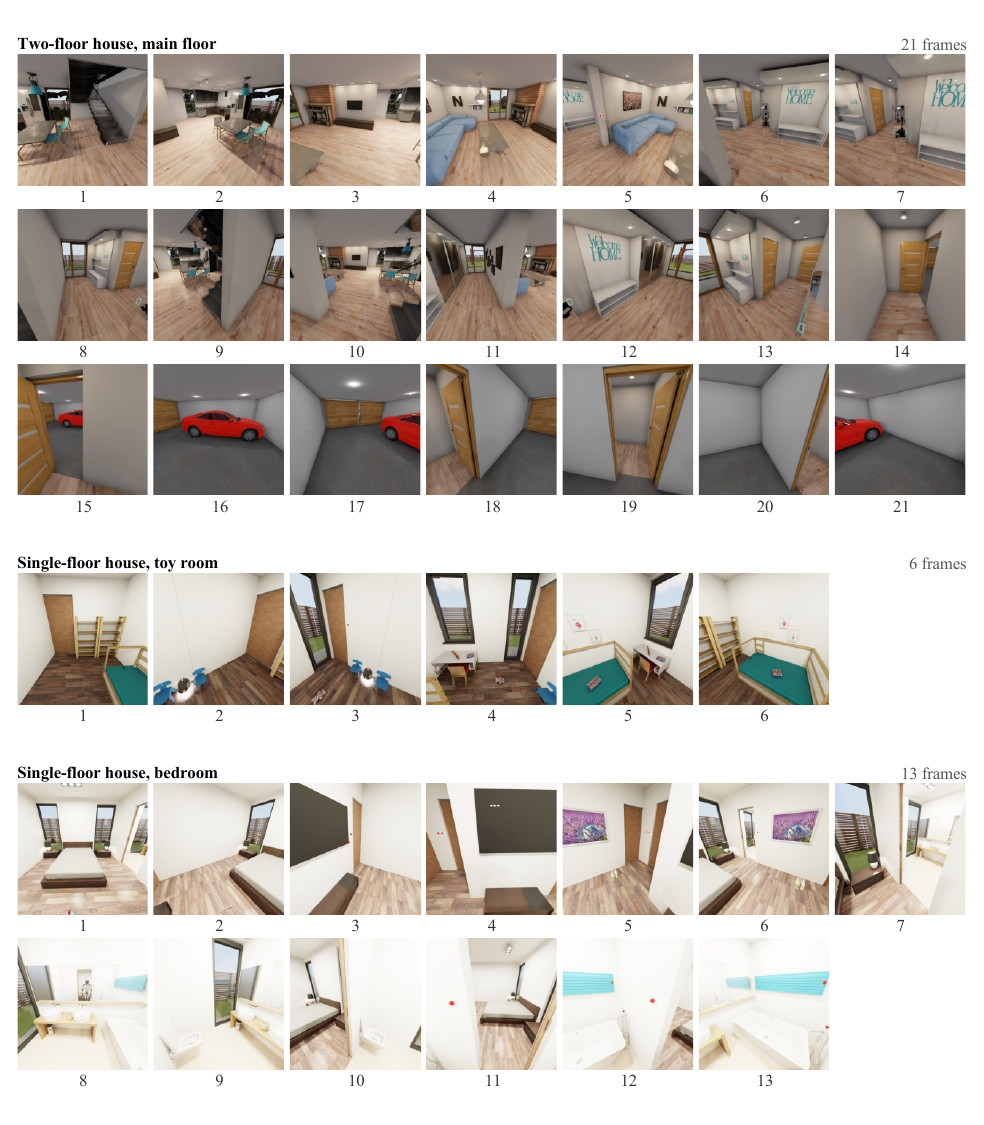}
    \vspace{-0.1cm}
    \caption{
    \textbf{Complete navigation context tours (1/2).}
    Context frames for Two-floor house, main floor;
    Single-floor house, toy room; and Single-floor house, bedroom.
    Within each scene, frames are arranged in input order,
    from left to right and then top to bottom.
    Numbers indicate one-based positions in the context sequence. The remaining scenes appear in Figure~\ref{fig:nav-context-part2}.
    }
    \vspace{-0.3cm}
    \label{fig:nav-context}
\end{figure*}

\begin{figure*}[p]
    \centering
    \includegraphics[width=0.9\textwidth]{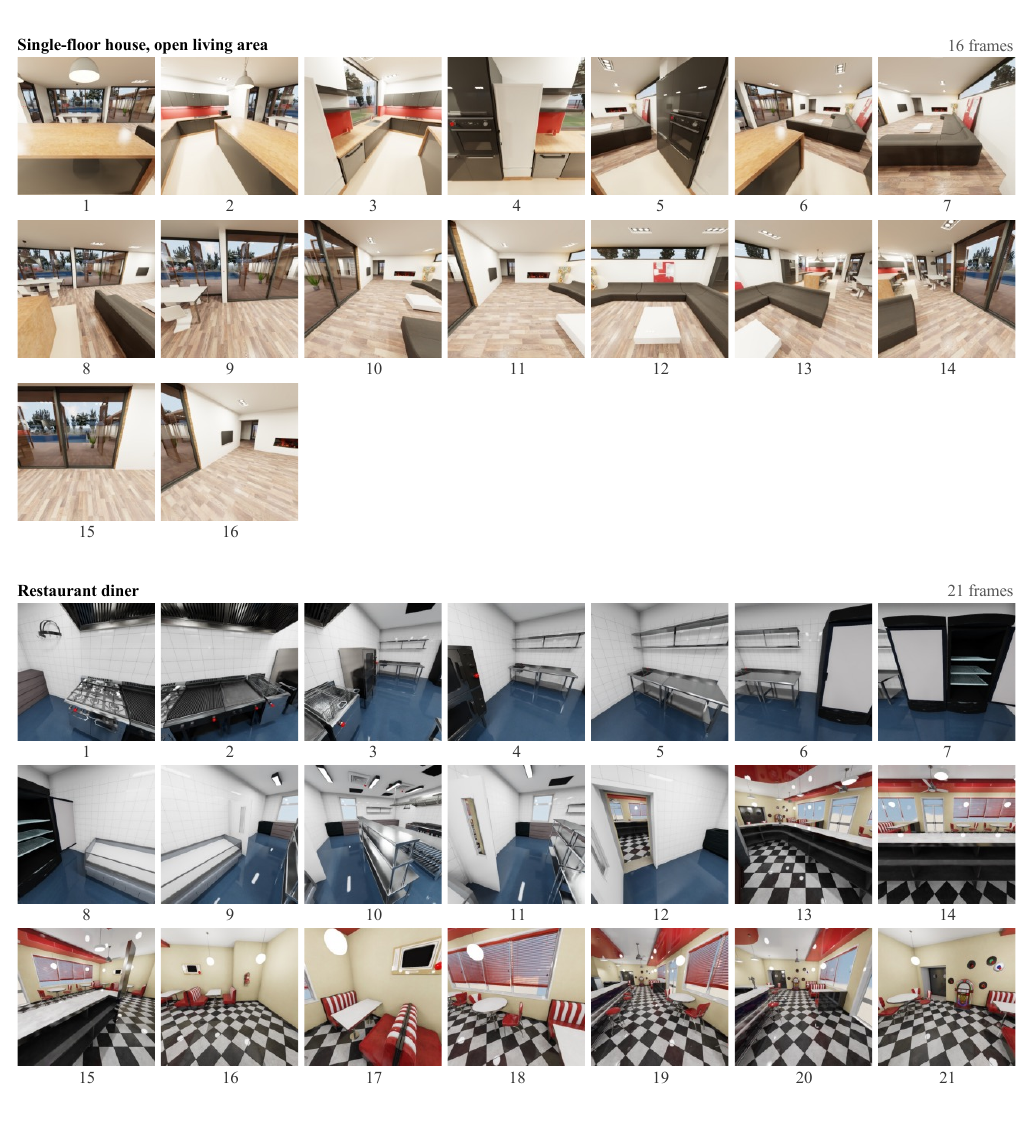}
    \vspace{-0.1cm}
    \caption{
    \textbf{Complete navigation context tours (2/2).}
    Context frames for Single-floor house, open living area;
    and Restaurant diner.
    Using the ordering and numbering
    convention of Figure~\ref{fig:nav-context}.
    }
    \vspace{-0.3cm}
    \label{fig:nav-context-part2}
\end{figure*}

%% file: supp_text/appx_model_details.tex
\section{Additional Model Training and Evaluation Details}
\label{app:model-details}

\subsection{Policy Adaptation and Initialization}
\label{app:policy-implementation}

\paragraph{Baseline.} The Qwen3.5-4B backbone (4.54B parameters) is frozen; the action expert and the action, state, and time projections (0.30B) are trained from scratch.

\paragraph{LoRA adaptation.} We add LoRA adapters (rank 16, $\alpha=16$, no dropout, $B$ initialized to zero) to every linear layer of the language model and vision
encoder. This adds 39M trainable parameters (0.9\% of the backbone), trained jointly with the baseline's trainable components.

\paragraph{DROID initialization.}
We initialize from step 160k of DROID pretraining
(frozen VLM, joint-position actions padded to 23 dimensions, horizon 32,
batch size 256, peak LR $3\times10^{-4}$), so the transferred weights reside in the action expert and projections. Finetuning is otherwise identical to the baseline.





\subsection{Training Setup}
\label{app:policy-training}

Table~\ref{tab:train-hparams} lists the training hyperparameters. All runs
use AdamW ($\beta=(0.9,0.95)$), no weight decay, gradient clipping at 1.0,
bf16 without EMA, 4k steps of linear warmup, an action horizon of 32, and
four GPUs. Manipulation tasks are sampled uniformly
from the 14 benchmark tasks plus one training-only saucer task; navigation
samples uniformly over its 10 tasks.
\input{supp_text/tab_training_hparams}

\input{supp_text/appx_oracle_ir}

\subsection{Policy Evaluation Instance Construction}
\label{app:eval-instances}

\paragraph{Navigation test instances.}
\label{app:nav-eval}
We construct held-out navigation instances by varying
the robot's initial position and orientation within
each environment. Candidate starting states undergo
simulator-based spawn validation, including checks
for excessive displacement, vertical motion, and tilt
during physics settling.
Figure~\ref{fig:nav-train-heldout} provides spatial
context for these examples by showing evaluation
starting poses, destinations, and surrounding furniture.
For the six house tasks, the maps additionally show
cached training-trajectory coverage.

\input{fig_text/fig_nav_test_dist}

%% file: supp_text/tab_training_hparams.tex
\begin{table}[htbp]
\caption{Policy training hyperparameters. Checkpoint is
the training step whose weights produce all reported results.}
\label{tab:train-hparams}
\begin{center}
\small
\begin{tabular}{lcccc}
\multicolumn{1}{c}{\bf Policy}
& \multicolumn{1}{c}{\bf Batch}
& \multicolumn{1}{c}{\bf Peak LR}
& \multicolumn{1}{c}{\bf LR schedule}
& \multicolumn{1}{c}{\bf Checkpoint}
\\ \hline \\
\multicolumn{5}{l}{\it Manipulation} \\[2pt]
$\pi_{0.5}$-Qwen (frozen VLM)   & 256 & $4\times10^{-4}$   & cosine & 100k \\
\quad + LoRA-VLM                & 128 & $2.8\times10^{-4}$ & cosine & 110k \\
\quad + DROID pretrain init     & 256 & $4\times10^{-4}$   & cosine & 90k  \\
\quad + IR variants (all)       & 256 & $4\times10^{-4}$   & cosine & 100k \\[4pt]
\multicolumn{5}{l}{\it Navigation} \\[2pt]
$\pi_{0.5}$-Qwen (frozen VLM)   & 512 & $4.2\times10^{-4}$ & cosine & 60k \\
\end{tabular}
\vspace{-1em}
\end{center}
\end{table}

%% file: supp_text/appx_oracle_ir.tex
\subsection{Oracle Intermediate Representation Construction}
\label{app:oracle-ir}

\paragraph{Target localization.}
The oracle IR policy variants receive simulator-derived
grounding images during both training and evaluation. For reaching tasks, the reference
tracks the target object's current 3D bounding box. For placement tasks,
we fix the box at the object's destination in the final demonstration
frame and reproject this fixed world-space reference through the current
head camera. Thus, the placement cue marks the goal location throughout
the episode. We project the box corners using the camera intrinsics and
extrinsics, clip the resulting 2D bounding box to the image boundaries,
and take its integer midpoint. We scale this coordinate to the
$224\times224$ policy image and round to the nearest pixel.
The reference is geometric and does not use an occlusion mask; an
occluded target can therefore retain a cue. 

\paragraph{Dot IR.}
We resize the head-camera RGB image to $224\times224$ before drawing
a filled red marker (RGB $(255,51,51)$) of radius 4 pixels with a
1-pixel white outline at the target coordinate. When no valid target
projection is available, the image contains no marker. The annotated
image is supplied as an additional input alongside the original head
and two wrist-camera images.

\paragraph{Gaussian attention-map IR.}
For a target coordinate $(c_x,c_y)$, we construct
\begin{equation}
    H(x,y)=\exp\!\left(
    -\frac{(x-c_x)^2+(y-c_y)^2}{2\sigma^2}
    \right),
    \qquad \sigma=8\ \text{pixels},
\end{equation}
on the same $224\times224$ grid. The Gaussian has unit peak and is
not normalized to sum to one. We quantize its values as
$q(x,y)=\operatorname{round}(255H(x,y))$ and map them to RGB using
a 256-entry analytic jet lookup table sampled over $[0,0.875]$.
This produces a dark-blue background (RGB $(0,0,128)$) and a red peak
(RGB $(255,0,0)$). An invalid projection produces a uniform dark-blue
image. The map is a standalone image without an RGB underlay and
occupies the same additional image-input slot as dot IR. See Figure~\ref{fig:ir-representations} for the difference.

\input{fig_text/fig_oracle_ir}

\paragraph{Temporal conditioning.}
The current-view cue is recomputed for each policy observation using
the current camera pose. Dot and Gaussian rendering use the same
respective parameters during training and evaluation. The first-frame
anchor variant additionally retains the initial dot-marked head image
and supplies this fixed image throughout the episode as a fifth input.


%% file: fig_text/fig_oracle_ir.tex
\begin{figure*}[htbp]
    \centering
    \includegraphics[width=0.7\textwidth]{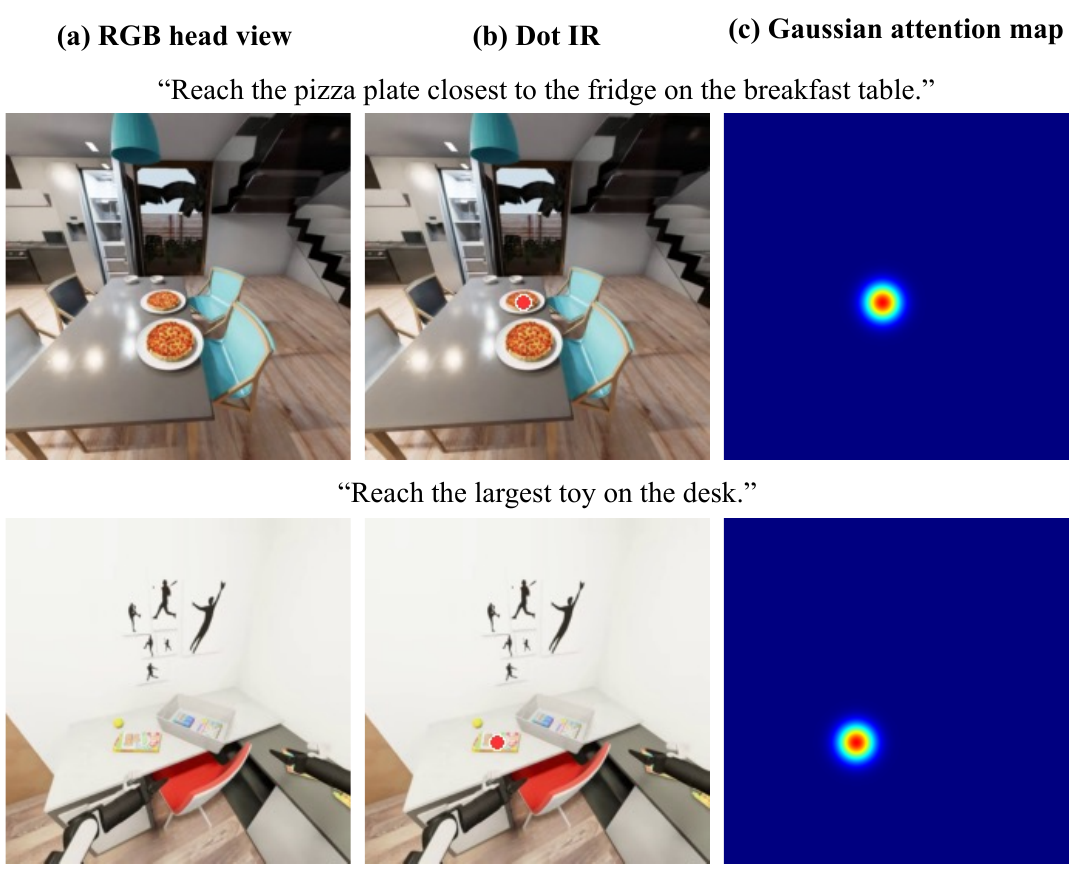}
    \caption{\textbf{Spatial intermediate representations.} Examples from distance (top) and size (bottom) tasks. Each row shows the RGB head-camera view, a dot IR marking the instructed target, and a Gaussian attention map centered at the same location.}
    \label{fig:ir-representations}
\end{figure*}

%% file: fig_text/fig_nav_test_dist.tex
\begin{figure*}[p]
    \centering
    \begin{minipage}[t]{0.6\textwidth}
        \vspace{0pt}
        \centering
        \includegraphics[width=\linewidth]
            {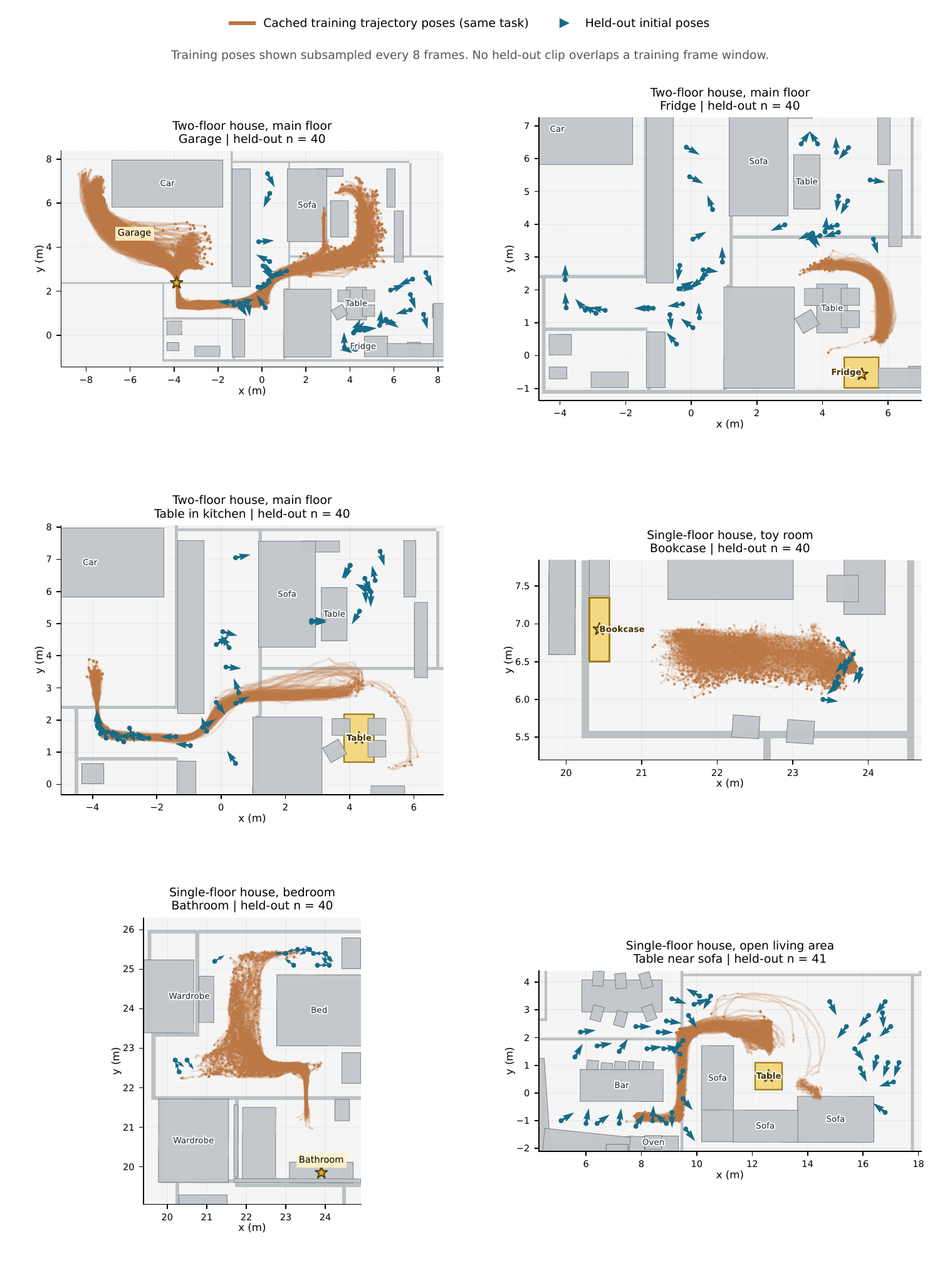}
        \par\small (a) Six house tasks
    \end{minipage}\hfill
    \begin{minipage}[t]{0.6\textwidth}
        \vspace{0pt}
        \centering
        \includegraphics[width=\linewidth]
            {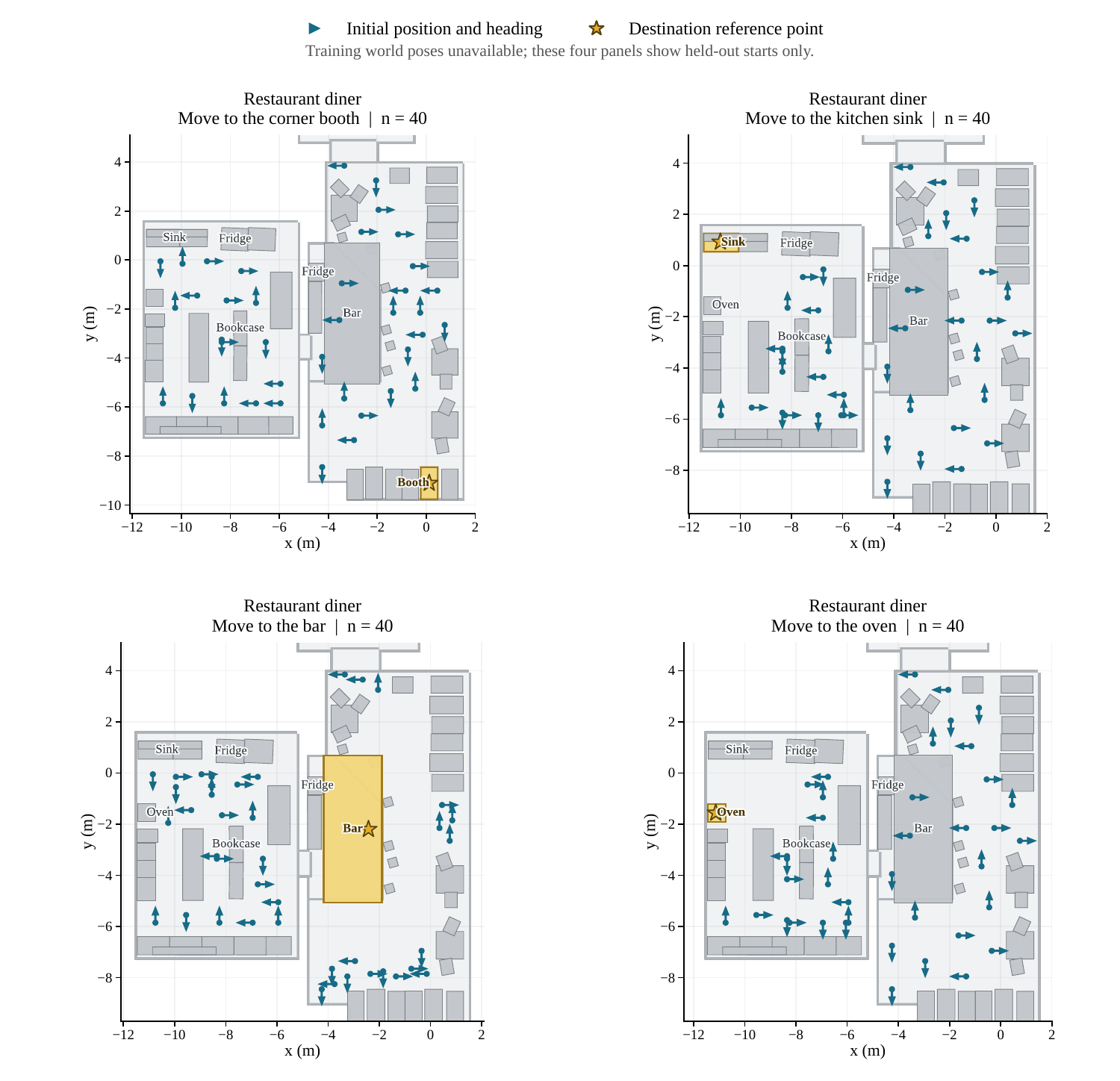}
        \par\small (b) Four diner tasks
    \end{minipage}

    \caption{
    \textbf{Navigation initial poses and scene layouts.}
    Blue arrows show evaluation starting positions and headings;
    orange points show cached training-trajectory poses for the
    corresponding house task. Training trajectories are unavailable
    for the diner tasks.
    Gray polygons show projected furniture bounding boxes,
    yellow polygons highlight target objects, and stars mark
    destination reference points.
    Bounding-box projections are schematic, not exact collision
    boundaries.
    }
    \label{fig:nav-train-heldout}
\end{figure*}

%% file: supp_text/appx_metrics.tex
\section{Evaluation Scoring Details}
\label{app:metrics}

\paragraph{VQA Aggregation.}
We report VQA accuracy within each spatial category and task.
Task-grounded manipulation VQA uses an unweighted mean of
per-task accuracies; navigation VQA is reported per task and
pooled over questions.

\paragraph{Reach IR Scoring.}
Predicted target points are evaluated against annotated target
regions derived from instance segmentation. Predictions assigned
to an accepted target region are labeled \emph{correct}, those
assigned to a non-target object are labeled \emph{wrong}, and
unassigned or unparseable predictions are labeled \emph{none}.
Tolerant scoring additionally accepts points within approximately
2\% of the image diagonal of the target region
(approximately 20 pixels for a $720\times720$ image), while
preserving all strictly correct predictions.
Tolerance expansion applies only to reach tasks. See Figure~\ref{fig:ir-scoring} (a) for illustration.

\input{fig_text/fig_ir_scoring}

\paragraph{Placement IR Scoring.}
For directional placement, the predicted point defines a camera
ray whose intersection with the countertop determines the
placement location. Its projection onto the robot-right axis
determines whether it lies on the instructed side of the reference.
Locations within $0.17$\,m of the reference center receive an
\emph{on-ref} outcome and no correct credit; off-surface predictions
are incorrect. For container placement, the ray is intersected
with the container volume, and its entry height is compared with
the instructed compartment's height band ($\pm0.05$\,m).
The reported manipulation IR average weights each of the 14 evaluated tasks equally. See Figure~\ref{fig:ir-scoring} (b) for illustration.

\paragraph{Manipulation Outcome Rules.}
Table~\ref{tab:success-criteria} specifies task-family completion
criteria, soft-credit conditions, and exceptions.
For non-committed reach episodes, soft credit requires the
end-effector's minimum distance to the instructed target over
the episode to be smaller than its minimum distance to every
distractor. For directional placement, the still-held object must
lie on the instructed side at episode end. For container placement,
the object's peak height while within the container footprint must
fall within the permitted compartment band; passing through the
band before overshooting does not qualify.
Floor-drop failures and released \emph{on-ref} outcomes receive
no additional soft credit.

The soda-can task additionally accepts base proximity within
$0.70$\,m under a nearest-candidate rule. Commitments to
non-instructed candidates or locations count as wrong-target
outcomes. Released \emph{on-ref} outcomes count as failures.
Soft credit captures observable target-consistent progress;
it does not directly establish a policy's internal grounding
decision.

\input{supp_text/appx_success_criteria}

%% file: fig_text/fig_ir_scoring.tex
\begin{figure*}[t]
\centering
\includegraphics[width=0.85\textwidth]
{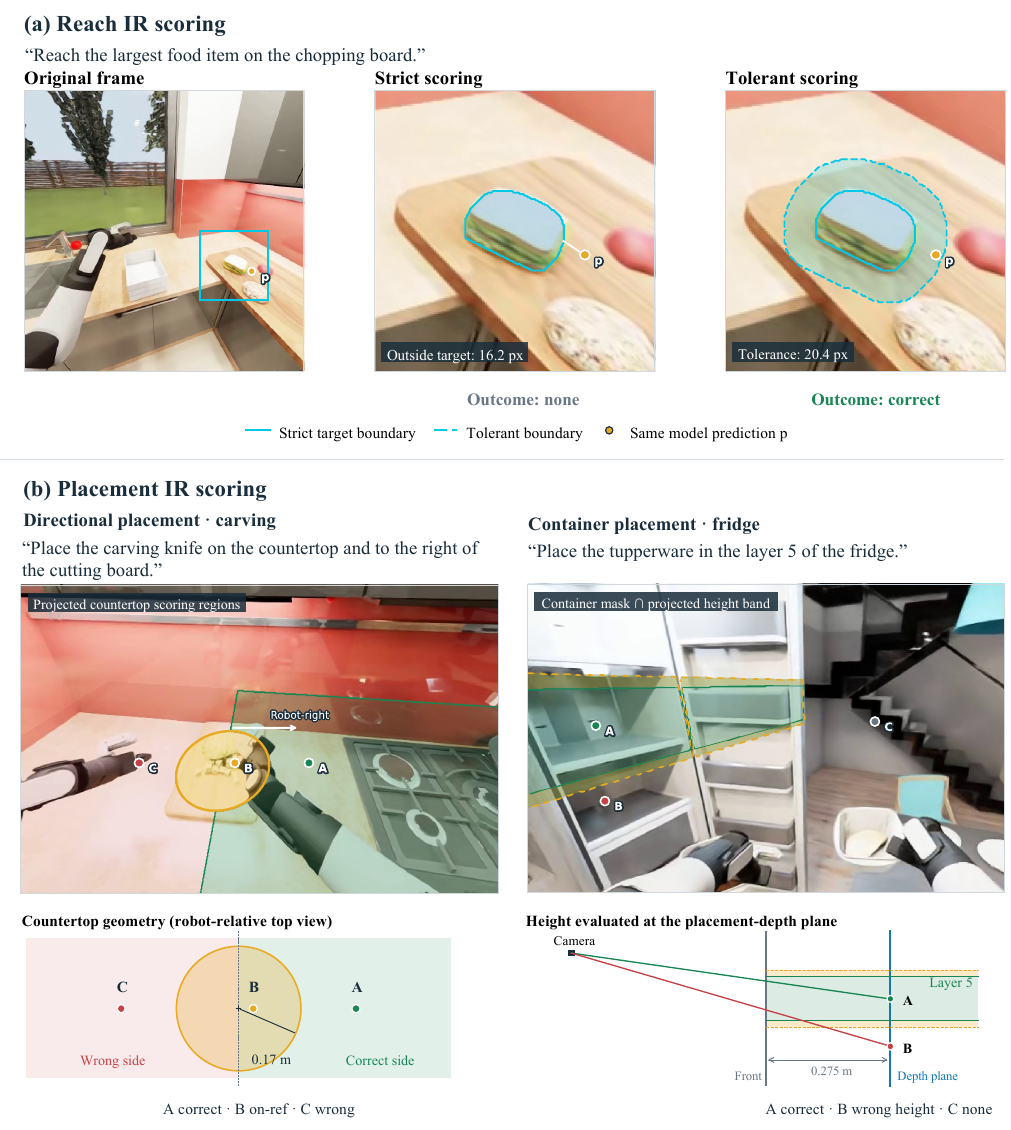}
\caption{\textbf{IR scoring on dataset frames.} (a) Strict and tolerant reach scoring for the same model prediction. (b) Directional and container placement scoring, with illustrative points A--C.}
\label{fig:ir-scoring}
\end{figure*}

%% file: supp_text/appx_success_criteria.tex
\begin{table}[htbp]
\caption{\textbf{Task-family success criteria.} Soft success includes strict
success plus eligible non-committed episodes; floor drops receive
no soft credit.}
\label{tab:success-criteria}
\begin{center}
\scriptsize
\setlength{\tabcolsep}{3pt}
\begin{tabular}{L{2.3cm}L{4.4cm}L{4.4cm}L{2.2cm}}
\multicolumn{1}{c}{\bf Family}
& \multicolumn{1}{c}{\bf Strict success (correct)}
& \multicolumn{1}{c}{\bf Additional soft credit (for none)}
& \multicolumn{1}{c}{\bf Tasks}
\\ \hline \\
Reach (contact) &
The end-effector touches the instructed candidate.
Touching a non-instructed candidate is \emph{wrong}. &
The minimum end-effector distance to the instructed target over
the episode is smaller than the minimum distance to every distractor. &
food, item, plate, toys, dice, paperback, pizza, storage, tool,
hardback \\[3pt]
Reach (base proximity) &
The end-effector touches the instructed can, or the base ends
within $0.70$\,m of it and satisfies the nearest-candidate rule. &
Same minimum-distance rule as contact reach. &
soda \\[3pt]
Directional placement &
The object is released onto the countertop, at least $0.17$\,m
from the reference, on the instructed side along the robot-right
axis. The opposite side is \emph{wrong}; releases within
$0.17$\,m or off the countertop are \emph{on-ref} failures. &
At episode end, the still-held object lies on the instructed
side along the robot-right axis. &
carving, saucer \\[3pt]
Container placement &
The object rests inside the instructed compartment's height band.
A different compartment is \emph{wrong}; outside the container
is \emph{none}. &
The object's \emph{peak} height within the container footprint
falls within the permitted band around the instructed compartment.
Passing through the band before overshooting does not qualify. &
tupperware, bowl \\[3pt]
Navigation &
The robot base satisfies the instructed goal criterion
(binary \texttt{subtask\_success}); no strict/soft split applies. &
--- &
all \texttt{move\_to} \\
\end{tabular}
\vspace{-2em}
\end{center}
\end{table}

%% file: supp_text/appx_vqa_eval.tex
\section{VQA and Point-Grounding Evaluation Prompts}
\label{app:eval-prompts}

This section documents the prompt components for
\vqaname and \IRname, including spatial conventions,
multi-frame context instructions, and output formats.
Text inside boxes reproduces the evaluation prompt
wording, with line wrapping adjusted for presentation.

\subsection{VQA Evaluation Prompts}
\label{app:vqa-prompts}

VQA prompts concatenate an observation-setting preamble,
an answer-format instruction, applicable spatial rules,
and the question with its answer choices.
Table~\ref{tab:vqa-templates} lists the general VQA
question templates.

\paragraph{Single-frame preamble.}
Single-frame questions begin with the following instruction.

\begin{center}
\fbox{\begin{minipage}{0.94\linewidth}
\scriptsize\ttfamily\raggedright
Answer the question based on the given image(s).
\end{minipage}}
\end{center}

\paragraph{Answer format.}
Multiple-choice questions receive the following
response-format instruction.

\begin{center}
\fbox{\begin{minipage}{0.94\linewidth}
\scriptsize\ttfamily\raggedright
Answer with the option's letter and text from the given
choices directly.\par
Example: if you think the correct answer is 'C.front'
from 'A.above B.under C. front D.behind.', your response
should only be 'C.front'.
\end{minipage}}
\end{center}

\paragraph{Direction rules.}
Non-locate direction questions receive the following
spatial conventions, except for task-grounded navigation,
which uses the robot-relative instructions below.

\begin{center}
\fbox{\begin{minipage}{0.94\linewidth}
\scriptsize\ttfamily\raggedright
Spatial Understanding Rules:\par
- Use the camera's perspective as the reference frame
for all spatial relationships (unless explicitly stated
otherwise).\par
- Front/Back: The object that appears farther from the
camera (deeper into the scene) is considered "in front of"
the object that appears closer to the camera. The object
closer to the camera is at the "back" of the farther
object.\par
- Left/Right: Defined from the camera's viewpoint.
An object located on the left side of the image relative
to another object is considered "left of" that object,
and similarly for right.\par
- Above/Below: Determined by relative vertical position
in the scene.
\end{minipage}}
\end{center}

\paragraph{Locate rules.}
Locate questions receive conventions for interpreting
container regions.

\begin{center}
\fbox{\begin{minipage}{0.94\linewidth}
\scriptsize\ttfamily\raggedright
Locate Question Rules:\par
- Grid Positions (Row x Column): Rows are counted from
bottom to top. Columns are counted from left to right.
For example, "1x1" is the bottom-left grid, "2x3" is the
second row from the bottom and third column from the
left.\par
- Layers: Counted from bottom to top of the container.
Layer 1 is the bottommost level (closest to the
floor/base), and higher layers are numbered sequentially
upward.\par
- Levels: Counted from bottom to top, described by their
relative height within the container. (e.g., "low level,"
"middle level," "high level," or specific numerical
levels)\par
- Please ONLY consider the internal space and
organizational structure of the specific container
mentioned in the question.
\end{minipage}}
\end{center}

\paragraph{Camera-distance rules.}
Distance questions referencing the camera receive the
following clarification.

\begin{center}
\fbox{\begin{minipage}{0.94\linewidth}
\scriptsize\ttfamily\raggedright
Camera Distance Question Rules:\par
- When questions refer to "the camera", note that the
robot hand visible in the image is NOT the camera.
The camera is the viewpoint from which the image was
captured, which is separate from any robot hand or
manipulator visible in the scene.\par
- Distance is defined along the view trace from the
camera to the 3Dcenter of the object.
\end{minipage}}
\end{center}

\paragraph{General multi-frame questions.}
For general multi-frame questions, the single-frame
preamble is replaced by the following instruction.

\begin{center}
\fbox{\begin{minipage}{0.94\linewidth}
\scriptsize\ttfamily\raggedright
These are consecutive observations from the head camera
of a robot. Answer the question based on the given images.
\end{minipage}}
\end{center}

Rotation questions additionally receive:

\begin{center}
\fbox{\begin{minipage}{0.94\linewidth}
\scriptsize\ttfamily\raggedright
The robot is rotating in place. The images show different
viewpoints as the robot turns.
\end{minipage}}
\end{center}

Navigation questions additionally receive:

\begin{center}
\fbox{\begin{minipage}{0.94\linewidth}
\scriptsize\ttfamily\raggedright
The robot is navigating through the environment.
The images show the robot's observations as it moves.
\end{minipage}}
\end{center}

\paragraph{Task-grounded navigation.}
The current task-grounded navigation prompt uses a
separate preamble that distinguishes context frames
from the current observation. It replaces the general
multi-frame preamble and direction-rule block.

\begin{center}
\fbox{\begin{minipage}{0.94\linewidth}
\scriptsize\ttfamily\raggedright
You are looking through the head camera of a mobile
robot.\par\smallskip
All images except the last one are a CONTEXT TOUR:
earlier views recorded while moving through this
environment. Use them to learn where things are.
They were NOT taken from the robot's current
position.\par\smallskip
The LAST image is the robot's CURRENT camera view.
All answers are relative to the robot's current position
and heading in that last image.\par\smallskip
Definitions (robot-centric, top-down):\par
- FRONT = the direction the camera is facing;
BACK = directly behind the robot.\par
- LEFT / RIGHT = the robot's left and right when facing
forward.\par
- The target may be OUT OF VIEW (behind the robot, or
behind a wall). It still has a direction. Locate the
target from the tour, locate the robot from the current
view, and answer with the quadrant the target lies in.
\end{minipage}}
\end{center}

\paragraph{Question and choices.}
The question and answer choices are appended after
the applicable instructions in the following format.

\begin{center}
\fbox{\begin{minipage}{0.94\linewidth}
\scriptsize\ttfamily\raggedright
Question: \{question\}\par\smallskip
Options: \{options\}
\end{minipage}}
\end{center}

\subsection{Point-Grounding Evaluation Prompts}
\label{app:ir-prompts}

For \IRname, the model receives the task instruction
and an image, and predicts one target point.
The prompt specifies integer coordinates normalized
to a 0--1000 scale and requires a JSON response.

\paragraph{Point-grounding template.}
\label{fig:ir_question_template}
\begin{center}
\fbox{\begin{minipage}{0.94\linewidth}
\scriptsize\ttfamily\raggedright
The robot's task is: \{instruction\}\par
\{layer\_hint\}\par
Predict the center of the target in the frame:
\{target description\}.\par
Output the spatial coordinate of the center as x, y
(integer coordinates normalized to a 0-1000 scale).\par
You must provide a valid prediction for the target ---
do not return null or empty values.\par
Predict exactly ONE point: x and y must each be a single
integer, never a list. If multiple objects are plausible,
commit to the single most likely one.\par
Respond ONLY with valid JSON in the exact format shown
below. Do not include any other text.\par\smallskip
Example: \{"results": \{"target":
\{"x": \textless int\textgreater,
"y": \textless int\textgreater\}\}\}
\end{minipage}}
\end{center}

\paragraph{Target description.}
For object targets, the target description is:

\begin{center}
\fbox{\begin{minipage}{0.94\linewidth}
\scriptsize\ttfamily\raggedright
the object that the robot must reach or manipulate
to accomplish this task
\end{minipage}}
\end{center}

For placement targets, it is:

\begin{center}
\fbox{\begin{minipage}{0.94\linewidth}
\scriptsize\ttfamily\raggedright
the target location where the robot must place the
object it is holding to accomplish this task
\end{minipage}}
\end{center}

\paragraph{Layer-counting hint.}
For placement instructions containing the word
``layer'', the following hint is inserted immediately
after the task instruction. Otherwise, the hint is
omitted.

\begin{center}
\fbox{\begin{minipage}{0.94\linewidth}
\scriptsize\ttfamily\raggedright
Layers are counted from bottom to top: for example,
"layer 5" is the fifth layer from the bottom of the fridge.
\end{minipage}}
\end{center}

%% file: supp_text/appx_experiment.tex
\section{Additional Experimental Analyses}
\label{app:experiment-details}

\subsection{Extracted VLA Backbones on Spatial VQA}
\label{app:backbone-vqa-diagnostic}
The extracted OpenHelix and $\pi_{0.5}$ backbones attain 9.6\% and
0.9\% single-frame accuracy under the tested VQA interface. Because
invalid answers are scored as incorrect, these scores combine answer
correctness with the ability to produce a valid option. 


\subsection{Qualitative Perception Examples}
\label{app:perception-examples}
Figure~\ref{fig:vqa_ir_samples} shows some examples for VQA and IR predictions on the same set of evaluated frames.
\input{fig_text/fig_vqa_ir_samples}

\subsection{Policy Execution and Checkpoint Selection}
\label{app:execution-settings}

\paragraph{Timing and action alignment.}
Each predicted chunk is indexed from the simulation step of its input
observation, so a chunk that arrives late is entered at a
latency-dependent offset. The controller keeps the committed chunk
while this index is below ten and within the prediction horizon;
otherwise, it switches to the newest chunk covering the current step.
Because the threshold counts from the observation rather than from
commitment, plan switches do not occur at a fixed interval.

\paragraph{Ensembling and fallback.}
Non-gripper dimensions, including the mobile base, are averaged over
the entries of a three-entry commitment history whose horizons cover
the current step. The weights are $w_r \propto \exp(0.01r)$, with
$r = 0,\ldots,m-1$ ordered oldest to newest and $m \le 3$; for $m=3$,
they are approximately $(0.330, 0.333, 0.337)$. Gripper commands
instead come from the newest chunk covering the current step, which
need not be the committed one. If no committed entry covers the
current step, the controller uses the most recently committed chunk
with its index clamped to the horizon. Before any chunk is committed,
it repeats the last action.

\paragraph{Checkpoint selection.}
The main manipulation comparison uses checkpoints at nearby, rather than identical, training steps: 100k updates for the baseline and the three core IR variants, 110k for LoRA-VLM, and 90k for the DROID-initialized policy.

\input{supp_text/appx_pertask_baseline}

\input{supp_text/appx_cap}

\subsection{Task-Level Correlation Plots}
\label{app:correlation-plots}
Each point in Figure~\ref{app:correlation} represents one manipulation task for a policy variant. The perception scores are from Qwen3.5-4B, and policy outcomes are soft held-out success rates. Lines show descriptive linear fits; the main Table~\ref{tab:perception_action_correlation} reports Spearman rank correlations, a distinct statistic.
\input{fig_text/fig_correlation_line_plots}

\subsection{Ground-truth vs.\ VLM-Predicted IR at Rollout}
\label{app:predicted-ir}

To test whether explicit target conditioning remains beneficial with predicted grounding, we replace the simulator-derived dot at inference with a VLM-predicted point while keeping the oracle-trained dot-IR policy fixed. The grounding VLM receives the current head-camera image at $512\times512$ resolution and the task instruction, and predicts the target object location for reaching or the destination for placement. The predicted point is rescaled to the $224\times224$ policy image and rendered using the same dot interface. A failed prediction produces an unmarked image; simulator coordinates are not used as a fallback. We evaluate Qwen3.5-4B, Qwen3.5-9B, and Molmo2-8B as grounding VLMs, with the same Qwen3.5-4B policy in all three cases.

\input{fig_text/tab_vlm_pred_ir_task_eval}

On the 469 held-out episodes, predicted dots improve soft success from 41.6\% to 47.1--48.8\% and strict success from 27.9\% to 33.5--34.5\% (Table~\ref{tab:vlm-predicted-ir}). These improvements recover approximately 29--39\% of the gain from oracle dots, which achieve 60.3\% soft and 44.8\% strict success. Thus, explicit target conditioning remains beneficial with VLM-predicted grounding, although a substantial gap to oracle conditioning remains.

The three grounding VLMs yield similar rollout success rates:
their results span 1.7 percentage points on soft success and
1.0 point on strict success, with no model leading on both metrics. These end-to-end results complement the IR grounding accuracy evaluated in the main paper: rollout success measures whether the predicted cues support successful control, rather than grounding accuracy alone.

\input{fig_text/fig_paperback_aug_ex}

\subsection{Training-Distribution Case Study}
\label{app:paperback-bias}

We study a separate \texttt{reach\_paperback\_book} training setup in which the referred book is associated with a fixed side in the demonstrations. To vary the approach geometry, we prefix each demonstration with motion from an initial robot position on the perpendicular bisector of the segment connecting the two candidate books to the original starting pose (Figure~\ref{fig:paperback-augmentation}).
The original interaction is retained.

We compare policies trained with and without this augmentation on the same 69 prompt-probe instances (Table~\ref{tab:paperback-aug}). Augmentation reduces wrong-target commitments from 79.7\% to 43.5\% and increases soft success from 11.6\% to 36.2\%, while strict success remains 10.1\%. These results
show that changing the demonstration distribution can improve
target selection and partial progress without increasing task
completion.

The main 14-task evaluation uses the augmented demonstrations for \texttt{reach\_paperback\_book}. Here, we isolate this task to compare training with and without augmentation on a shared prompt-probe set.
The augmentation changes motion trajectories and data exposure
along with the spatial regularities of the demonstrations, so
the observed effect cannot be attributed solely to reduced
side bias.
\input{fig_text/tab_aug_data_bias_res}

\subsection{Data Scaling}
\label{app:datascaling}

We study how demonstration volume affects the dot-IR policy
by retraining it on nested 30\% and 50\% subsets of
\datasetname and comparing against the full-data model.
Subsets are sampled per task to preserve the task mixture.
All models are trained for the same number of updates and
evaluated under the same protocol.
Table~\ref{tab:main-results} reports aggregate results, and
Tables~\ref{tab:pertask-selected} and~\ref{tab:pertask-probe}
provide per-task breakdowns.

The full-data model achieves the highest pooled soft and strict
success on both evaluation sets. From 30\% to 100\% data,
held-out soft success increases from 56.1\% to 60.3\%
($+4.2$ percentage points), while strict success increases
from 34.5\% to 44.8\% ($+10.3$ points).
Prompt-probe soft and strict success also improve, from
32.6\% to 38.0\% ($+5.4$ points) and from 19.9\% to 24.9\%
($+5.0$ points), respectively. The prompt-probe wrong-object
commitment rate decreases from 46.6\% to 43.9\%.
Thus, additional demonstrations improve both success and
object selection in the aggregate; these results do not
isolate the benefit to execution alone.

\input{supp_text/tab_pertask_data_scaling}

Scaling is not monotonic at the intermediate data fraction.
At 50\% data, held-out soft/strict success is 53.7\%/33.7\%,
compared with 56.1\%/34.5\% at 30\%; prompt-probe success
is also slightly lower at 50\% (30.6\%/19.4\%) than at 30\%
(32.6\%/19.9\%). Nevertheless, the 30\% dot-IR policy exceeds the full-data frozen-VLM baseline by 14.5 and 6.6 points in held-out soft and strict success, and by 16.5 and 9.9 points on prompt-probe.

Per-task results reveal substantial variation.
Held-out strict success on \texttt{reach\_tool} increases
from 15.6\% at 30\% data to 37.5\% at 100\%, and
\texttt{place\_carving\_knife} improves from 12.5\% to 33.3\%.
In contrast, held-out \texttt{reach\_storage\_box} changes
little (65.6\% to 68.8\% strict), although its prompt-probe
strict success rises from 30.0\% to 52.0\%.
More data does not help every task:
\texttt{place\_bowl} performs best at reduced data fractions
on both evaluation sets. These differences indicate that
the benefit of additional demonstrations depends on both
the task and the evaluation setting.

\subsection{Per-Task Policy Comparisons}
\label{app:pertask}

Tables~\ref{tab:pertask-selected} and~\ref{tab:pertask-probe} provide
per-task held-out and prompt-probe results for the frozen-VLM baseline and dot IR policies used in
\S\ref{app:datascaling}. The 100\%, 50\%, and 30\% columns show
the effect of training-data fraction for that configuration.

\subsection{Navigation Results}\label{app:navigation-results} Table~\ref{tab_nav-results} reports VQA accuracy and navigation success for all ten tasks, with the scene or region specified for each task.\par\input{fig_text/tab_nav_result}\subsection{Qualitative Policy and Navigation-VQA Examples}
\label{app:qualitative-policy}

The qualitative navigation examples illustrate potential sources of
difficulty beyond the aggregate results in
Section~\ref{sec:exp-nav}. Longer routes expose the policy to more
obstacles, while narrow doorway crossings require precise base motion.
Observed failures include collisions with scene objects and unsuccessful
doorway traversal, suggesting that route geometry contributes to the variation in task success.

Tasks requiring travel opposite to the order of the context frames
(fridge, kitchen sink, and oven) also have relatively low success.
This pattern is consistent with reliance on the demonstrated traversal
direction, although differences in route geometry prevent us from
isolating the effect of context order. A related probe in the two-floor
house found success in 19/20 episodes starting in the trained region
and 0/20 episodes starting from novel kitchen locations. These results
suggest limited generalization across starting locations, but do not
establish whether the limitation arises from context interpretation,
route familiarity, or motion execution.

Figures~\ref{fig:vla-samples}, \ref{fig:nav_samples},
and~\ref{fig:nav_vqa} show selected manipulation-policy,
navigation-policy, and navigation-VQA examples, respectively.

\input{fig_text/fig_vla_samples}
\input{fig_text/fig_nav_samples}
\input{fig_text/fig_nav_vqa}

%% file: fig_text/fig_vqa_ir_samples.tex
\begin{figure*}[ht]
    \centering
    \includegraphics[width=0.97\textwidth]{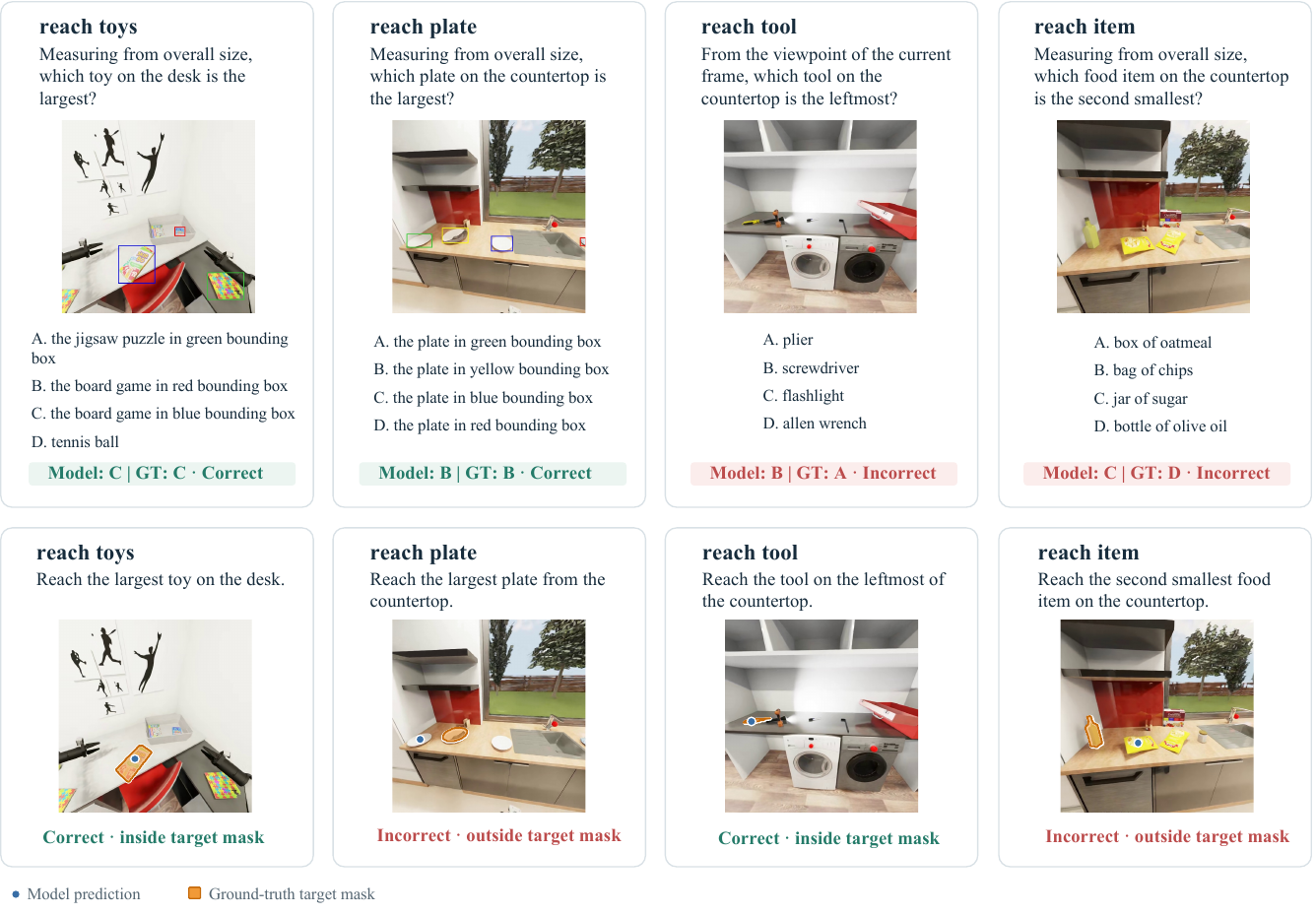}
    \vspace{-0.1cm}
    \caption{\textbf{Qualitative examples of VQA and IR predictions.} Qwen3.5-4B predictions on held-out task frames, with VQA answers above and IR predictions below. Columns show both correct, VQA-only correct, IR-only correct, and both incorrect. Blue points indicate IR predictions; orange masks mark reference targets. IR is correct only when the predicted point falls inside the target mask. Questions and options are reproduced verbatim; images are cropped for readability.}
    \vspace{-0.3cm}
    \label{fig:vqa_ir_samples}
\end{figure*}

%% file: supp_text/appx_pertask_baseline.tex
\subsection{Per-Task Baseline Results}
\label{app:failure-analysis}

Table~\ref{tab:heldout-covariates} provides the per-task held-out
results supporting the baseline analysis in \S\ref{sec:experiments}.
Strict success, soft success, and wrong-target rates distinguish
completed interactions, target-consistent partial progress, and
incorrect target commitments. Task statistics are included for context.

\begin{table}[htbp]
\caption{Per-task held-out results of the frozen-VLM baseline with
covariates. Strict/Soft = strict/soft success (\%); Wrong =
wrong-target commitments (\%); Cand.\ = mean same-category candidates
per scene; Size = mean apparent size of the target object (square root
of its 2D bounding-box area in the $720^2$ head-camera evaluation
frames, px); Frames = training frames (k); Len.\ = mean demonstration
length (frames). A dash indicates that candidate count is not reported
for that task. All rates use the task's $n$ evaluated episodes.}
\label{tab:heldout-covariates}
\begin{center}
\small
\begin{tabular}{lcccccccc}
\multicolumn{1}{c}{\bf Task} & \multicolumn{1}{c}{\bf $n$}
& \multicolumn{1}{c}{\bf Strict$\uparrow$}
& \multicolumn{1}{c}{\bf Soft$\uparrow$}
& \multicolumn{1}{c}{\bf Wrong$\downarrow$}
& \multicolumn{1}{c}{\bf Cand.}
& \multicolumn{1}{c}{\bf Size (px)}
& \multicolumn{1}{c}{\bf Frames (k)}
& \multicolumn{1}{c}{\bf Len.}
\\ \hline \\
reach\_food\_item             & 32 & 53.1 & 56.2 & 34.4 & 4.0 &  52 & 319 &  318 \\
reach\_storage\_box           & 32 & 46.9 & 68.8 & 25.0 & 2.0 &  85 & 145 & 1365 \\
reach\_soda\_can              & 42 & 38.1 & 52.4 & 42.9 & 3.0 &  35 &  97 &  696 \\
reach\_plate                  & 31 & 35.5 & 35.5 & 61.3 & 3.7 & 124 &  55 &  212 \\
reach\_hardback               & 48 & 31.2 & 37.5 & 37.5 & 4.3 &  66 & 237 &  421 \\
reach\_item                   & 32 & 31.2 & 34.4 & 53.1 & 7.0 &  98 & 307 &  311 \\
place\_carving\_knife         & 24 & 29.2 & 70.8 &  4.2 & --- &  62 &  58 &  184 \\
reach\_paperback\_book        & 22 & 27.3 & 54.5 & 18.2 & 2.0 &  80 & 193 &  602 \\
reach\_toys                   & 40 & 22.5 & 30.0 & 37.5 & 5.8 &  70 & 126 &  441 \\
reach\_pizza\_plate           & 40 & 20.0 & 50.0 & 25.0 & 2.0 &  94 & 288 & 1212 \\
reach\_dice                   & 32 & 18.8 & 56.2 &  6.2 & 2.0 &  22 & 145 &  793 \\
place\_bowl\_in\_cabinet      & 40 & 15.0 & 15.0 & 77.5 & --- & 110 & 202 &  450 \\
place\_tupperware\_in\_fridge & 22 & 13.6 & 13.6 & 45.5 & --- & 132 & 200 &  501 \\
reach\_tool                   & 32 &  6.2 & 15.6 & 46.9 & 4.6 &  53 &  87 &  461 \\
\end{tabular}
\vspace{-2em}
\end{center}
\end{table}

%% file: supp_text/appx_cap.tex
\subsection{Code-as-Policy Agent Evaluation}
\label{app:cap}

This appendix supplements \S\ref{cap-agents} and \S\ref{sec:cap-mani} with
program-development details, execution settings, and
per-task results. The agent interface and evaluation metrics
are described in the main paper.

\subsubsection{Program Development and Selection}

\paragraph{Writer context.}
For each task, all writers receive an identical system prompt
containing the API reference, restrictions on privileged
information, development instructions, task-specific scoring
notes, and spatial-relation conventions.
Each request includes the last three text exchanges.
The initial request additionally includes head- and wrist-camera
images from an example scene. Later requests include the
previous episode's verdict, printed output, API call log,
final camera views, and a history of development outcomes.
Each response contains a short plan and a complete Python
module defining \texttt{run(api)}.

\paragraph{Development episodes.}
Each writer develops its program over 10 training episodes
per task. Feedback includes the strict outcome and whether
the hand ended nearer the correct or a wrong candidate.
This simulator-derived feedback is available during development,
not during test execution.

Before simulation, candidate programs are screened using a mock
API constructed from a recorded scene. Each development step
allows up to three writing attempts. Programs are rejected if they raise an error, never command a motion, issue 100 consecutive perception calls without a motion, or make no API call for 20 s; programs still running at the 60 s or 500-call cap are accepted. If all three attempts fail, the previous version (if it passed the check) is run instead. If the previous version also failed, the step is skipped and the writer is told no episode was run. At the first step, a failing program is still run. These checks are applied during development only.

\paragraph{Validation and freezing.}
After the 10 development episodes, the final program is run
once on a held-aside episode from the training distribution.
This episode is distinct from the development episodes,
test episodes, and example scene.
Validation requires strict success; target-consistent partial
progress alone does not pass.

If the final program passes validation, it is frozen.
Otherwise, the most recent version that achieved strict success
during development is frozen. If no development version
succeeded, the final program is frozen despite failing validation.


\subsubsection{Writer Implementation}

Qwen3.5-4B and Qwen3.5-9B use
\texttt{Qwen/Qwen3.5-4B} and \texttt{Qwen/Qwen3.5-9B},
served with vLLM 0.19.1 in \texttt{bf16}.
Both use thinking mode with the \texttt{qwen3} reasoning
parser, temperature 1.0, top-$p$ 0.95, top-$k$ 20,
min-$p$ 0, and presence penalty 1.5.

Gemini 3 uses \texttt{gemini-3.1-pro-preview}
through Vertex AI with high thinking level.
GPT-6 uses \texttt{gpt-6-astra} through the OpenAI
Responses API with high reasoning effort.
Fable 5.1 uses \texttt{claude-fable-5-1} through the
Anthropic Messages API with adaptive thinking,
high effort. 
Other generation settings use the respective API defaults.

\subsubsection{Agent API}
\label{app:cap-api}

Programs define \texttt{run(api)} and interact with the robot only through the
\texttt{api} object below. Positions are in metres in the robot base frame
($x$ forward, $y$ left, $z$ up, origin on the floor under the robot); the frame moves
with the robot, so positions must be re-expressed or re-observed after base motion.
Cameras are \texttt{head}, \texttt{left\_wrist}, and \texttt{right\_wrist}.
Perception and state calls do not advance simulated time; motion, gripper, and
\texttt{wait} calls block until completion. Programs may import \texttt{numpy},
\texttt{math}, \texttt{re}, \texttt{cv2}, \texttt{PIL}, \texttt{scipy}, and the Python
standard library.

\paragraph{Observation and state.}
\begin{itemize}
  \item \texttt{instruction}: the natural-language instruction for the episode.
  \item \texttt{get\_state()}: end-effector poses of both arms, gripper opening, grasp
        state, joint positions, and the current step.
  \item \texttt{get\_image(camera)}: the camera's RGB image ($H{\times}W{\times}3$).
  \item \texttt{get\_depth(camera)}: the camera's depth image (metres along the optical axis,
        pixel-aligned with the RGB image).
\end{itemize}

\paragraph{Geometry.}
\begin{itemize}
  \item \texttt{pixel\_to\_point(u, v, camera)}: the 3-D base-frame point seen at pixel
        $(u,v)$, from the camera's depth (median of a $5{\times}5$ patch).
  \item \texttt{point\_to\_pixel(p, camera)}: the pixel at which base-frame point $p$
        appears, or none if behind the camera.
  \item \texttt{pixel\_to\_ray(u, v, camera)}: the ray (origin, unit direction) through
        pixel $(u,v)$.
\end{itemize}

\paragraph{Perception tools.}
\begin{itemize}
  \item \texttt{point\_at(text, camera)}: image points for every instance of \texttt{text}
        in the current image, from an open-vocabulary pointing model
        (Molmo2-8B
        ; prompt ``Point to every \{text\} in the image.'', greedy
        decoding). \texttt{text} may be any description, including the full instruction.
  \item \texttt{ask\_vlm(question, cameras)}: a free-form answer from a VLM (GPT-5, low reasoning effort) about the current images of the given cameras.
\end{itemize}

\paragraph{Motion and gripper.}
\begin{itemize}
  \item \texttt{approach(p, standoff, arm)}: turns the base to face point $p$ and drives
        until $p$ is \texttt{standoff} metres ahead (default 0.55), offset toward the
        reaching arm; returns $p$ re-expressed in the new base frame.
  \item \texttt{move\_base(dx, dy, dyaw)}: relative base motion; stops short if blocked.
  \item \texttt{move\_ee\_to(p, arm, speed, use\_torso)}: straight-line end-effector motion to
        point $p$ (orientation not controlled); \texttt{use\_torso} lets the torso lean for
        extra reach.
  \item \texttt{move\_ee\_by(d, arm, speed, use\_torso)}: relative end-effector motion.
  \item \texttt{open\_gripper(arm)}, \texttt{close\_gripper(arm)}: gripper control.
  \item \texttt{wait(seconds)}: hold still.
\end{itemize}
Every motion call returns its success and, on failure, a reason
(\texttt{stalled}: contact or reach limit; \texttt{blocked}: base obstructed;
\texttt{pushing}: excessive force, after which the arm backs off; \texttt{timeout}).
With \texttt{arm="auto"}, the arm already holding an object is used; otherwise the left
arm for targets with $y \geq 0$ and the right arm otherwise.

\paragraph{Utilities.}
\begin{itemize}
  \item \texttt{log(msg, **data)}: write to the episode log. Printed output is returned to
        the writer during development.
\end{itemize}


\subsubsection{Execution and Evaluation}

\paragraph{Closed-loop execution.}
Frozen programs are executed without writer revisions on all
test episodes of their task. They can repeatedly observe,
query perception tools, inspect motion results, branch,
and retry actions. Thus, freezing the program does not
prevent closed-loop behavior.

The simulator is paused during computation, perception calls,
and state queries. Only motion and gripper commands and \texttt{wait} advance simulated time; these calls block until
completion. Evaluation imposes no explicit tool-call limit.
The perception tools retry network errors up to four times.
If a program issues no robot command for 900 seconds of
wall-clock time, it is stopped and the robot holds still
until the episode ends.


\paragraph{Simulation budgets and termination.}
The nominal simulation budgets match those used for the
VLA policies. Writers may declare completion, but termination and success are determined by the simulator under the same step budgets as the VLA policies.

\paragraph{Metrics.}
We evaluate all writers on the same held-out set across 14 manipulation tasks, using the task-specific
strict-success, soft-success, and wrong-target criteria
defined in \S\ref{sec:metrics}.
Table~\ref{tab:e2-per-task} reports per-task soft and strict
success.

\input{fig_text/tab_e2_per_task}



%% file: fig_text/tab_e2_per_task.tex
\begin{table}[htbp]
\caption{\textbf{Per-task code-as-policy performance.}
Each cell reports soft / strict success (\%).
$N$ is the number of test episodes.
Bold indicates the highest soft success within each task,
including ties.
Per-task values are rounded to whole percentages;
the pooled row is computed from aggregate episode counts.}
\label{tab:e2-per-task}
\centering
\footnotesize
\setlength{\tabcolsep}{3pt}
\resizebox{\linewidth}{!}{%
\begin{tabular}{@{}lrccccc@{}}
\multicolumn{1}{c}{\bf Task}
& {\bf $N$}
& \multicolumn{3}{c}{\bf Proprietary writers}
& \multicolumn{2}{c}{\bf Open-weight writers} \\
& & Fable 5.1 & GPT-6 & Gemini 3
& Qwen3.5-9B & Qwen3.5-4B
\\ \hline
\texttt{place\_bowl\_in\_cabinet}
& 40 & {\bf 72} / 72 & 48 / 32 & 28 / 28 & 20 / 20 & 18 / 3 \\
\texttt{place\_carving\_knife}
& 24 & {\bf 100} / 58 & 75 / 58 & 38 / 21 & 50 / 8 & 71 / 0 \\
\texttt{place\_tupperware\_in\_fridge}
& 22 & 18 / 18 & {\bf 59} / 9 & 5 / 5 & 18 / 9 & 5 / 5 \\
\texttt{reach\_dice}
& 32 & {\bf 97} / 94 & 91 / 88 & 78 / 56 & 66 / 0 & 56 / 0 \\
\texttt{reach\_food\_item}
& 32 & {\bf 69} / 59 & 59 / 56 & 59 / 53 & 47 / 6 & 44 / 6 \\
\texttt{reach\_hardback\_from\_bookcase}
& 48 & {\bf 83} / 48 & 42 / 19 & 27 / 19 & 21 / 0 & 19 / 0 \\
\texttt{reach\_item}
& 32 & {\bf 56} / 50 & 53 / 50 & 34 / 28 & 3 / 0 & 9 / 0 \\
\texttt{reach\_paperback\_book\_aug}
& 22 & 64 / 45 & {\bf 73} / 64 & 64 / 41 & 50 / 0 & 45 / 0 \\
\texttt{reach\_pizza\_plate}
& 40 & 55 / 22 & {\bf 72} / 28 & 58 / 12 & 50 / 0 & 62 / 0 \\
\texttt{reach\_plate}
& 31 & 45 / 35 & {\bf 65} / 61 & 45 / 32 & 32 / 0 & 32 / 0 \\
\texttt{reach\_soda\_can}
& 42 & 64 / 57 & {\bf 76} / 76 & 64 / 45 & 57 / 7 & 52 / 26 \\
\texttt{reach\_storage\_box}
& 32 & 72 / 59 & 78 / 72 & {\bf 88} / 66 & 56 / 25 & 47 / 28 \\
\texttt{reach\_tool}
& 32 & 56 / 41 & 47 / 38 & 56 / 3 & {\bf 62} / 0 & 31 / 0 \\
\texttt{reach\_toys}
& 40 & {\bf 60} / 50 & 38 / 28 & 58 / 52 & 22 / 0 & 32 / 0 \\
\hline
{\bf Pooled}
& 469
& {\bf 66.1} / 51.4
& 61.2 / 47.3
& 50.3 / 33.3
& 39.0 / 5.3
& 37.1 / 5.1 \\
\end{tabular}%
\vspace{-2em}
}
\end{table}

%% file: fig_text/fig_correlation_line_plots.tex
\begin{figure*}[htbp]
    \centering
    \includegraphics[width=0.95\textwidth]{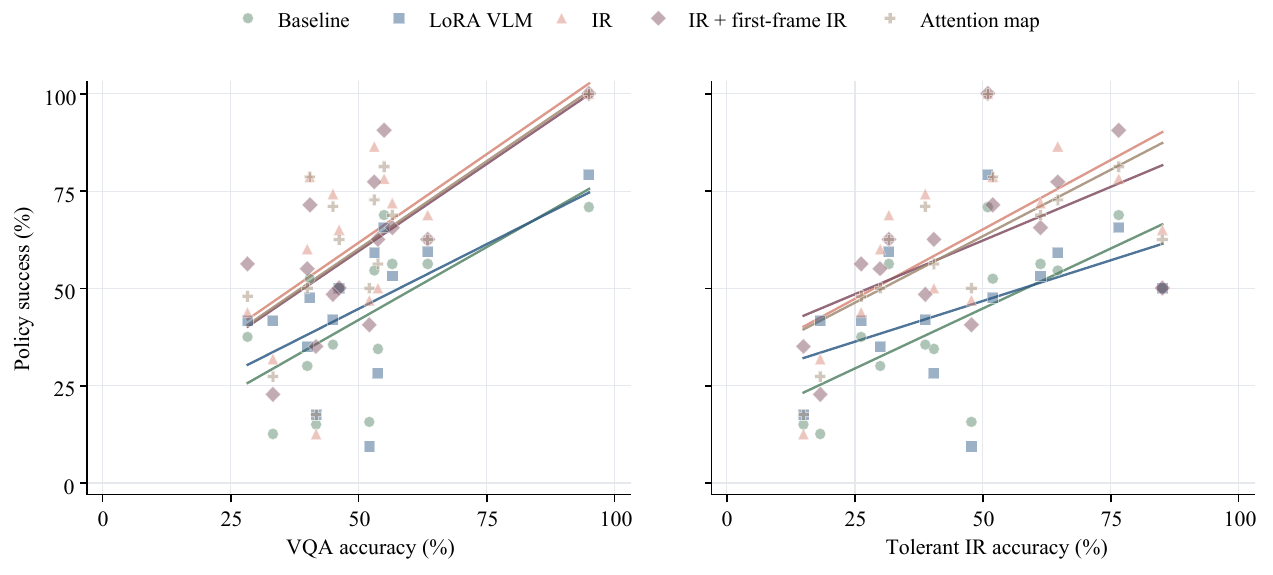}
    \vspace{-0.1cm}
    \caption{\textbf{Task-level correlation between perception and policy success.} Each point represents a manipulation task for a policy variant, pairing a Qwen3.5-4B perception score with the policy's soft held-out success rate. Lines show descriptive linear fits.}
    \vspace{-0.3cm}
    \label{app:correlation}
\end{figure*}

%% file: fig_text/tab_vlm_pred_ir_task_eval.tex
\begin{table}[htbp]
\caption{\textbf{Rollout with VLM-predicted dot IR.}
Success (\%) on the same held-out set.
All rows use the Qwen3.5-4B policy backbone; predicted rows replace the
ground-truth dot IR at inference with a dot predicted by the listed
grounding VLM.}
\label{tab:vlm-predicted-ir}
\begin{center}
\small
\begin{tabular}{lcc}
\multicolumn{1}{c}{\bf IR source}
& \multicolumn{1}{c}{\bf Soft$\uparrow$}
& \multicolumn{1}{c}{\bf Strict$\uparrow$}
\\ \hline \\
Baseline                    & 41.6 & 27.9 \\[2pt]
\multicolumn{3}{l}{\it Predicted dot} \\[2pt]
\quad Qwen3.5-4B            & 48.8 & 34.1 \\
\quad Qwen3.5-9B            & 47.3 & 34.5 \\
\quad Molmo2-8B             & 47.1 & 33.5 \\[2pt]
Oracle dot                  & 60.3 & 44.8 \\
\end{tabular}
\end{center}
\end{table}

%% file: fig_text/fig_paperback_aug_ex.tex
\begin{figure*}[htbp]
    \centering
    \includegraphics[width=0.8\textwidth]{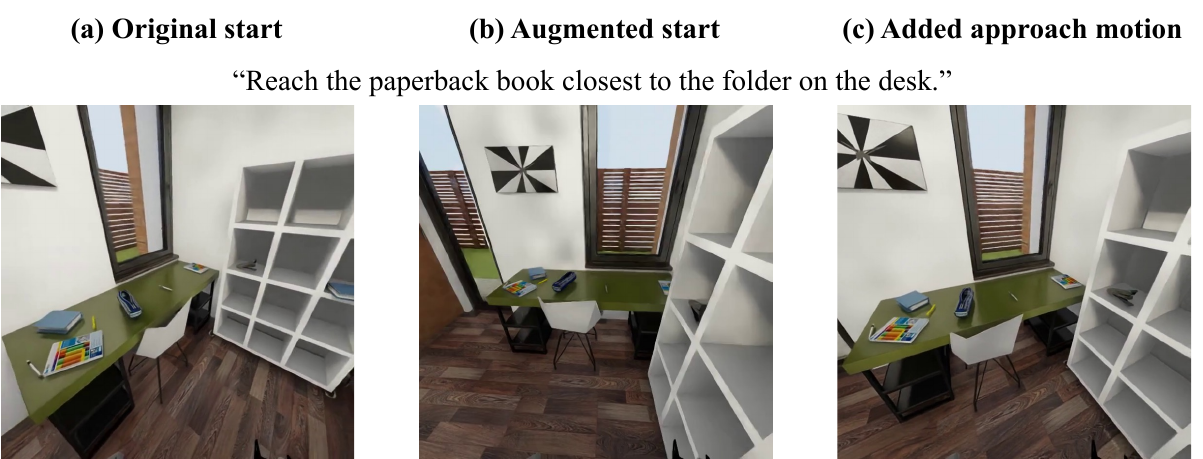}
    \caption{\textbf{Original and augmented paperback demonstrations.}
    Matched observations show (a) the original starting pose,
    (b) the augmented starting pose, and (c) an intermediate frame
    during the added approach motion. The augmented demonstration
    proceeds from (b) through (c) to (a), then follows the original
    interaction. The instruction and target remain unchanged.
    All panels are recorded head-camera frames from the same
    source demonstration and its augmented variant.}
    \label{fig:paperback-augmentation}
\end{figure*}

%% file: fig_text/tab_aug_data_bias_res.tex
\vspace{-1em}
\begin{table}[htbp]
\caption{\textbf{Paperback demonstration-augmentation case study results.}
Policies trained on original demonstrations or with approach-motion
prefix augmentation are evaluated on the same 69 prompt-probe
instances. All values are percentages; Wrong denotes wrong-target
commitments.}
\label{tab:paperback-aug}
\begin{center}
\small
\begin{tabular}{lccc}
\multicolumn{1}{c}{\bf Training data}
& \multicolumn{3}{c}{\bf Prompt-probe} \\
\multicolumn{1}{c}{}
& Soft$\uparrow$ & Strict$\uparrow$ & Wrong$\downarrow$
\\ \hline \\
Original                & 11.6 & 10.1 & 79.7 \\
+ prefix augmentation   & 36.2 & 10.1 & 43.5 \\
\end{tabular}
\vspace{-3em}
\end{center}
\end{table}

%% file: supp_text/tab_pertask_data_scaling.tex
\vspace{-1em}
\begin{table}[htbp]
\caption{Per-task \textbf{held-out} strict/soft success (\%).
Base = frozen-VLM baseline; Dot IR = dot-conditioned policy, at the indicated training-data fraction.
Bold marks the highest value in each row for each metric separately,
including ties. Pooled rates aggregate evaluated episodes.}
\label{tab:pertask-selected}
\begin{center}
\small
\begin{tabular}{lcccc}
\multicolumn{1}{c}{\bf Task}
& \multicolumn{1}{c}{\bf Base}
& \multicolumn{3}{c}{\bf Dot IR} \\
\multicolumn{1}{c}{} & & 100\% & 50\% & 30\%
\\ \hline \\
reach\_food\_item  & 53.1/56.2 & {\bf 65.6}/{\bf 71.9} & {\bf 65.6}/68.8 & 59.4/68.8 \\
reach\_plate  & 35.5/35.5 & {\bf 71.0}/{\bf 74.2} & 48.4/64.5 & 51.6/67.7 \\
reach\_soda\_can  & 38.1/52.4 & 71.4/78.6 & 69.0/{\bf 81.0} & {\bf 73.8}/{\bf 81.0} \\
reach\_storage\_box  & 46.9/68.8 & {\bf 68.8}/{\bf 78.1} & 65.6/{\bf 78.1} & 65.6/{\bf 78.1} \\
reach\_item  & 31.2/34.4 & {\bf 50.0}/{\bf 50.0} & {\bf 50.0}/{\bf 50.0} & 43.8/43.8 \\
reach\_paperback\_book  & 27.3/54.5 & {\bf 54.5}/{\bf 86.4} & 27.3/68.2 & 36.4/59.1 \\
reach\_pizza\_plate  & 20.0/50.0 & {\bf 32.5}/{\bf 65.0} & 17.5/52.5 & 20.0/62.5 \\
reach\_tool  & 6.2/15.6 & {\bf 37.5}/{\bf 46.9} & 15.6/34.4 & 15.6/37.5 \\
reach\_toys  & 22.5/30.0 & {\bf 42.5}/{\bf 60.0} & 25.0/37.5 & 25.0/55.0 \\
reach\_hardback  & {\bf 31.2}/37.5 & 29.2/{\bf 43.8} & 14.6/35.4 & 16.7/{\bf 43.8} \\
reach\_dice  & 18.8/56.2 & {\bf 37.5}/{\bf 68.8} & 15.6/59.4 & 25.0/62.5 \\
place\_carving\_knife  & 29.2/70.8 & {\bf 33.3}/{\bf 100.0} & 25.0/{\bf 100.0} & 12.5/79.2 \\
place\_tupperware  & 13.6/13.6 & {\bf 31.8}/{\bf 31.8} & 22.7/27.3 & 18.2/22.7 \\
place\_bowl  & 15.0/15.0 & 10.0/12.5 & 12.5/17.5 & {\bf 17.5}/{\bf 25.0} \\
\hline \\
\multicolumn{1}{l}{\bf Pooled}  & 27.9/41.6 & {\bf 44.8}/{\bf 60.3} & 33.7/53.7 & 34.5/56.1 \\
\end{tabular}
\end{center}
\end{table}

\begin{table}[htbp]
\caption{Per-task \textbf{prompt-probe} strict/soft success (\%)
for the same configurations as Table~\ref{tab:pertask-selected}.
Bold marks the highest value in each row for each metric separately,
including ties.}
\label{tab:pertask-probe}
\begin{center}
\small
\begin{tabular}{lcccc}
\multicolumn{1}{c}{\bf Task}
& \multicolumn{1}{c}{\bf Base}
& \multicolumn{3}{c}{\bf Dot IR} \\
\multicolumn{1}{c}{} & & 100\% & 50\% & 30\%
\\ \hline \\
reach\_food\_item  & 16.0/16.0 & {\bf 40.0}/{\bf 40.0} & 28.0/30.0 & 22.0/24.0 \\
reach\_soda\_can  & 12.0/16.0 & 62.0/62.0 & {\bf 76.0}/{\bf 76.0} & 66.0/68.0 \\
reach\_storage\_box  & 4.0/4.0 & {\bf 52.0}/{\bf 54.0} & 26.0/28.0 & 30.0/34.0 \\
reach\_plate  & 6.0/6.0 & 20.0/24.0 & 8.0/12.0 & {\bf 26.0}/{\bf 28.0} \\
reach\_tool  & 8.0/8.0 & {\bf 28.0}/32.0 & 22.0/{\bf 36.0} & 26.0/32.0 \\
reach\_paperback\_book  & 12.0/40.0 & {\bf 32.0}/{\bf 66.0} & 30.0/46.0 & 22.0/46.0 \\
reach\_dice  & 18.0/28.0 & {\bf 26.0}/{\bf 66.0} & 14.0/52.0 & 18.0/{\bf 66.0} \\
reach\_hardback  & {\bf 24.0}/24.0 & 20.0/{\bf 36.0} & 4.0/18.0 & 4.0/26.0 \\
reach\_item  & 4.0/4.0 & {\bf 14.0}/{\bf 14.0} & 8.0/10.0 & 12.0/12.0 \\
reach\_pizza\_plate  & {\bf 10.0}/26.0 & 2.0/{\bf 36.0} & 0.0/20.0 & 4.0/16.0 \\
reach\_toys  & 0.0/0.0 & {\bf 6.0}/{\bf 16.0} & 2.0/6.0 & 2.0/12.0 \\
place\_carving\_knife  & 8.0/26.0 & 20.0/56.0 & {\bf 22.0}/{\bf 58.0} & 12.0/50.0 \\
place\_tupperware  & 10.0/20.0 & {\bf 22.0}/{\bf 24.0} & 14.0/16.0 & 18.0/22.0 \\
place\_bowl  & 8.0/8.0 & 4.0/6.0 & {\bf 18.0}/{\bf 20.0} & 16.0/{\bf 20.0} \\
\hline \\
\multicolumn{1}{l}{\bf Pooled}  & 10.0/16.1 & {\bf 24.9}/{\bf 38.0} & 19.4/30.6 & 19.9/32.6 \\
\end{tabular}
\end{center}
\end{table}

%% file: fig_text/tab_nav_result.tex
\begin{table}[!h]
\caption{\textbf{Navigation performance across 10 tasks.}
VQA = Qwen3.5-4B accuracy on scene-based direction questions;
Success = \vlaname{} navigation success on spawn-validated
held-out instances. All values are percentages.}
\label{tab_nav-results}
\begin{center}
\small
\begin{tabular}{llcc}
\multicolumn{1}{c}{\bf Task}
& \multicolumn{1}{c}{\bf Scene / region}
& \multicolumn{1}{c}{\bf VQA$\uparrow$}
& \multicolumn{1}{c}{\bf Success$\uparrow$}
\\ \hline \\
move\_to\_bookcase  & single-floor house, toy room   & 30.0 & {\bf 92.5} \\
move\_to\_bathroom   & single-floor house, bedroom    & {\bf 40.0} & 90.0 \\
move\_to\_corner\_booth        & restaurant diner                           & 23.3 & 87.5 \\
move\_to\_table\_in\_kitchen   & two-floor house, main floor       & 38.2 & 75.0 \\
move\_to\_table\_near\_sofa    & single-floor house, open living area & 23.9 & 73.2 \\
move\_to\_bar                  & restaurant diner                           & 25.0 & 50.0 \\
move\_to\_kitchen\_sink        & restaurant diner                           & 16.7 & 45.0 \\
move\_to\_fridge               & two-floor house, main floor       & 33.8 & 35.0 \\
move\_to\_oven                 & restaurant diner                           & 36.7 & 27.5 \\
move\_to\_garage               & two-floor house, main floor       & 34.2 & 22.5 \\
\hline \\
\multicolumn{2}{l}{\bf Pooled} & 30.2 & 59.9 \\
\end{tabular}
\vspace{-2em}
\end{center}
\end{table}

%% file: fig_text/fig_vla_samples.tex
\begin{figure*}[ht]
    \centering
    \includegraphics[width=0.9\textwidth]{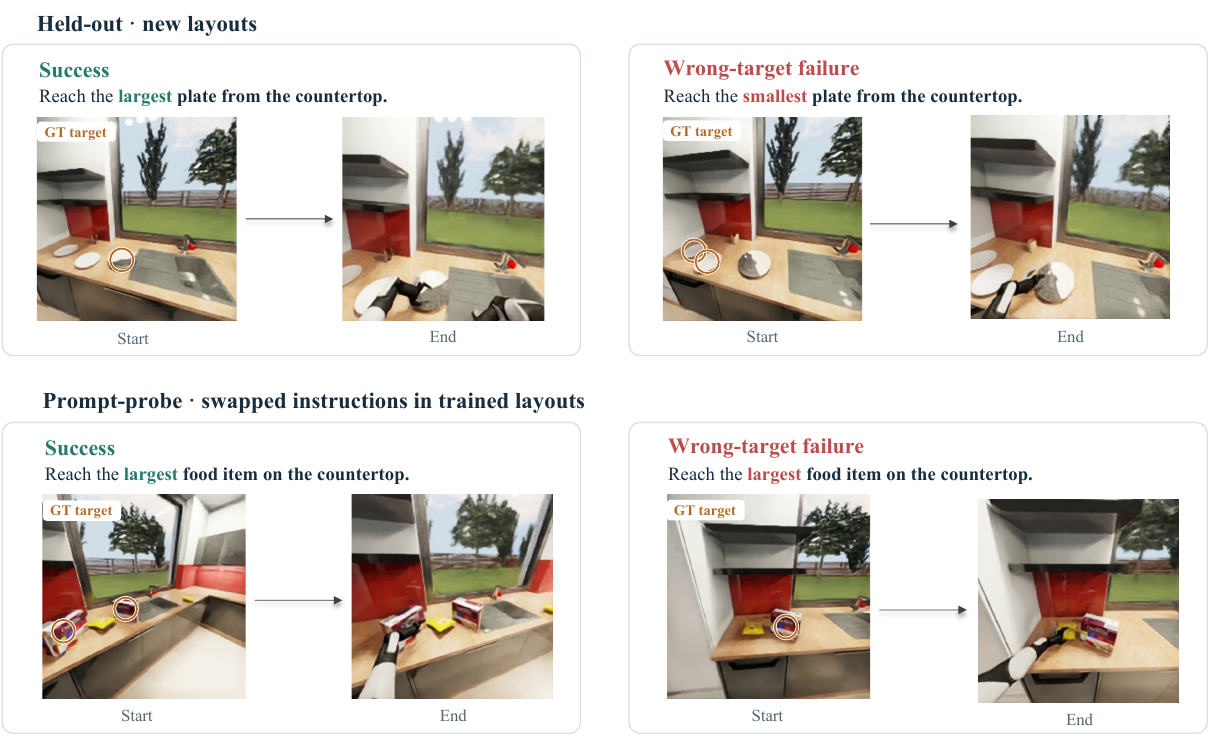}
    \vspace{-0.1cm}
    \caption{\textbf{Qualitative examples of VLA generalization and instruction following.} Start and end frames from four independent baseline evaluations. Rows show held-out layouts and swapped instructions in training layouts; columns show successes and wrong-target failures. Orange circles mark acceptable targets. In the failed instruction-swap probe, the policy reaches for the original training target (chip bag) instead of the instructed largest item (oatmeal box).}
    \vspace{-0.3cm}
    \label{fig:vla-samples}
\end{figure*}

%% file: fig_text/fig_nav_samples.tex
\begin{figure*}[ht]
    \centering
    \includegraphics[width=0.9\textwidth]{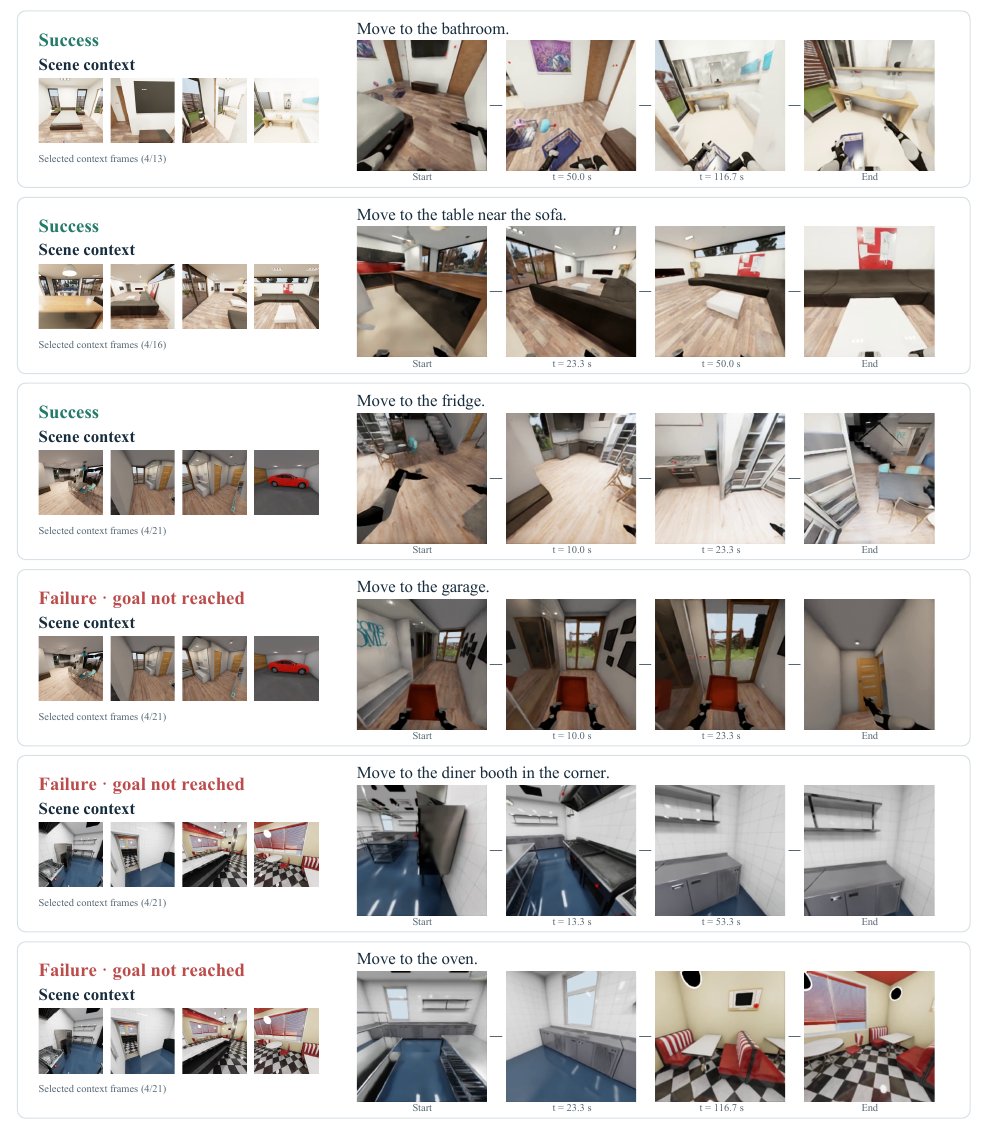}
    \vspace{-0.1cm}
    \caption{\textbf{Qualitative examples of navigation outcomes.}
    Each row shows selected scene-context and rollout
    frames from a held-out episode using the 10-task
    policy. The top three episodes
    succeed; the bottom three fail to reach their
    instructed goals.}
    \vspace{-0.3cm}
    \label{fig:nav_samples}
\end{figure*}

%% file: fig_text/fig_nav_vqa.tex
\begin{figure*}[ht]
    \centering
    \includegraphics[width=0.9\textwidth]{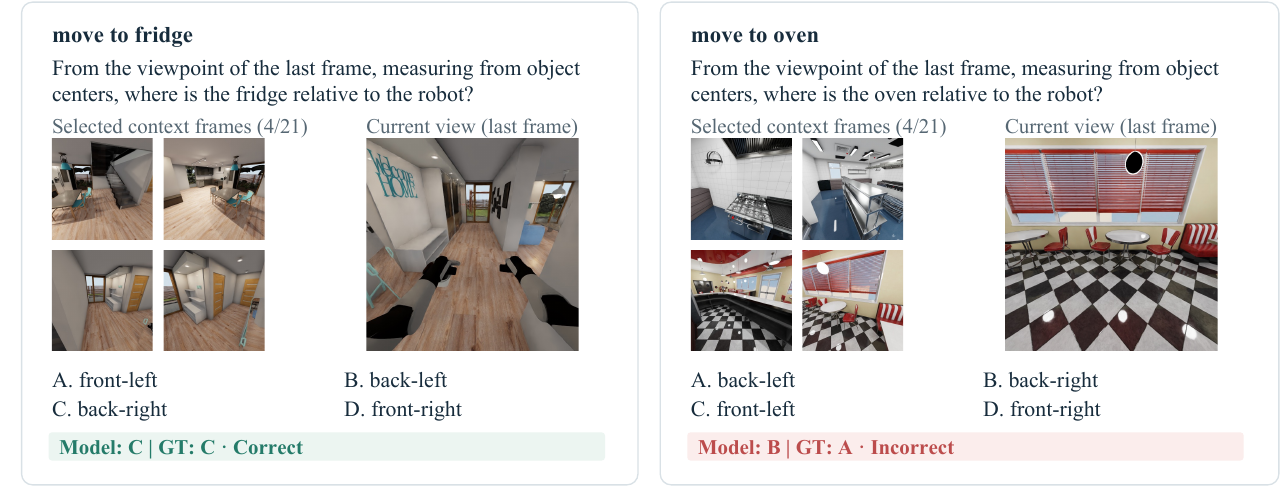}
    \vspace{-0.1cm}
    \caption{\textbf{Qualitative examples of navigation VQA predictions.} Correct and incorrect Qwen3.5-4B answers on held-out navigation questions. Each example shows four selected context frames and the current view, which is the last input frame referenced by the question.}
    \vspace{-0.3cm}
    \label{fig:nav_vqa}
\end{figure*}

%% file: iclr2027_conference.bib
@misc{qwen3.5-4b,
  author = {Qwen Team},
  title = {Qwen3.5-4B: A Unified Vision-Language Foundation Model},
  year = {2026},
  publisher = {Hugging Face},
  journal = {Hugging Face Repository},
  howpublished = {\url{https://huggingface.co/Qwen/Qwen3.5-4B}}
}

@article{li2024behavior1k,
    title   = {{BEHAVIOR-1K}: A Human-Centered, Embodied {AI} Benchmark with 1,000 Everyday Activities and Realistic Simulation},
    author  = {Chengshu Li and Ruohan Zhang and Josiah Wong and Cem Gokmen and Sanjana Srivastava and Roberto Martín-Martín and Chen Wang and Gabrael Levine and Wensi Ai and Benjamin Martinez and Hang Yin and Michael Lingelbach and Minjune Hwang and Ayano Hiranaka and Sujay Garlanka and Arman Aydin and Sharon Lee and Jiankai Sun and Mona Anvari and Manasi Sharma and Dhruva Bansal and Samuel Hunter and Kyu-Young Kim and Alan Lou and Caleb R Matthews and Ivan Villa-Renteria and Jerry Huayang Tang and Claire Tang and Fei Xia and Yunzhu Li and Silvio Savarese and Hyowon Gweon and C. Karen Liu and Jiajun Wu and Li Fei-Fei},
    journal = {arXiv preprint arXiv:2403.09227},
    year    = {2024}
}

@misc{openai2026gpt6astra,
  author       = {{OpenAI}},
  title        = {{GPT-6 Astra System Card}},
  year         = {2026},
  month        = sep,
  url          = {https://deploymentsafety.openai.com/gpt-6-astra},
  note         = {Accessed: 2026-09-26}
}

@misc{google2025gemini3pro,
  title        = {Gemini 3 Pro Model Card},
  author       = {{Google DeepMind}},
  year         = {2025},
  howpublished = {\url{https://deepmind.google/models/gemini/pro/}},
  note         = {Model \texttt{gemini-3-pro-preview}, accessed April 2026}
}

@misc{intelligence2025pi05visionlanguageactionmodelopenworld,
      title={$\pi_{0.5}$: a Vision-Language-Action Model with Open-World Generalization}, 
      author={Physical Intelligence and Kevin Black and Noah Brown and James Darpinian and Karan Dhabalia and Danny Driess and Adnan Esmail and Michael Equi and Chelsea Finn and Niccolo Fusai and Manuel Y. Galliker and Dibya Ghosh and Lachy Groom and Karol Hausman and Brian Ichter and Szymon Jakubczak and Tim Jones and Liyiming Ke and Devin LeBlanc and Sergey Levine and Adrian Li-Bell and Mohith Mothukuri and Suraj Nair and Karl Pertsch and Allen Z. Ren and Lucy Xiaoyang Shi and Laura Smith and Jost Tobias Springenberg and Kyle Stachowicz and James Tanner and Quan Vuong and Homer Walke and Anna Walling and Haohuan Wang and Lili Yu and Ury Zhilinsky},
      year={2025},
      eprint={2504.16054},
      archivePrefix={arXiv},
      primaryClass={cs.LG},
      url={https://arxiv.org/abs/2504.16054}, 
}

@misc{chen2024spatialvlmendowingvisionlanguagemodels,
      title={SpatialVLM: Endowing Vision-Language Models with Spatial Reasoning Capabilities}, 
      author={Boyuan Chen and Zhuo Xu and Sean Kirmani and Brian Ichter and Danny Driess and Pete Florence and Dorsa Sadigh and Leonidas Guibas and Fei Xia},
      year={2024},
      eprint={2401.12168},
      archivePrefix={arXiv},
      primaryClass={cs.CV},
      url={https://arxiv.org/abs/2401.12168}, 
}

@misc{cheng2024spatialrgptgroundedspatialreasoning,
      title={SpatialRGPT: Grounded Spatial Reasoning in Vision Language Models}, 
      author={An-Chieh Cheng and Hongxu Yin and Yang Fu and Qiushan Guo and Ruihan Yang and Jan Kautz and Xiaolong Wang and Sifei Liu},
      year={2024},
      eprint={2406.01584},
      archivePrefix={arXiv},
      primaryClass={cs.CV},
      url={https://arxiv.org/abs/2406.01584}, 
}

@misc{kamath2023whatsupvisionlanguagemodels,
      title={What's "up" with vision-language models? Investigating their struggle with spatial reasoning}, 
      author={Amita Kamath and Jack Hessel and Kai-Wei Chang},
      year={2023},
      eprint={2310.19785},
      archivePrefix={arXiv},
      primaryClass={cs.CL},
      url={https://arxiv.org/abs/2310.19785}, 
}

@misc{fu2024blinkmultimodallargelanguage,
      title={BLINK: Multimodal Large Language Models Can See but Not Perceive}, 
      author={Xingyu Fu and Yushi Hu and Bangzheng Li and Yu Feng and Haoyu Wang and Xudong Lin and Dan Roth and Noah A. Smith and Wei-Chiu Ma and Ranjay Krishna},
      year={2024},
      eprint={2404.12390},
      archivePrefix={arXiv},
      primaryClass={cs.CV},
      url={https://arxiv.org/abs/2404.12390}, 
}

@misc{yang2025thinkingspacemultimodallarge,
      title={Thinking in Space: How Multimodal Large Language Models See, Remember, and Recall Spaces}, 
      author={Jihan Yang and Shusheng Yang and Anjali W. Gupta and Rilyn Han and Li Fei-Fei and Saining Xie},
      year={2025},
      eprint={2412.14171},
      archivePrefix={arXiv},
      primaryClass={cs.CV},
      url={https://arxiv.org/abs/2412.14171}, 
}

@misc{wang2026mindcubespatialmentalmodeling,
      title={MindCube: Spatial Mental Modeling from Limited Views}, 
      author={Qineng Wang and Baiqiao Yin and Pingyue Zhang and Jianshu Zhang and Kangrui Wang and Zihan Wang and Jieyu Zhang and Keshigeyan Chandrasegaran and Han Liu and Ranjay Krishna and Saining Xie and Jiajun Wu and Li Fei-Fei and Manling Li},
      year={2026},
      eprint={2506.21458},
      archivePrefix={arXiv},
      primaryClass={cs.AI},
      url={https://arxiv.org/abs/2506.21458}, 
}

@misc{qi2026dissectingembodiedabilitiesmultimodal,
      title={Dissecting Embodied Abilities in Multimodal Language Models through Skill-level Evaluation and Diagnosis}, 
      author={Yu Qi and Haibo Zhao and Ziyu Guo and Siyuan Ma and Ziyan Chen and Yaokun Han and Renrui Zhang and Zitiantao Lin and Yizhe Zhu and Shiji Xin and Yijian Huang and Boce Hu and Kai Cheng and Peiheng Wang and Jiazheng Liu and Jiayi Zhang and Yizhe Zhu and Wenqing Wang and Yiran Qin and Haojie Huang and Lawson L. S. Wong},
      year={2026},
      eprint={2510.08759},
      archivePrefix={arXiv},
      primaryClass={cs.CV},
      url={https://arxiv.org/abs/2510.08759}, 
}

@misc{chen2025robo2vlmvisualquestionanswering,
      title={Robo2VLM: Visual Question Answering from Large-Scale In-the-Wild Robot Manipulation Datasets}, 
      author={Kaiyuan Chen and Shuangyu Xie and Zehan Ma and Pannag R Sanketi and Ken Goldberg},
      year={2025},
      eprint={2505.15517},
      archivePrefix={arXiv},
      primaryClass={cs.RO},
      url={https://arxiv.org/abs/2505.15517}, 
}

@misc{mees2022calvinbenchmarklanguageconditionedpolicy,
      title={CALVIN: A Benchmark for Language-Conditioned Policy Learning for Long-Horizon Robot Manipulation Tasks}, 
      author={Oier Mees and Lukas Hermann and Erick Rosete-Beas and Wolfram Burgard},
      year={2022},
      eprint={2112.03227},
      archivePrefix={arXiv},
      primaryClass={cs.RO},
      url={https://arxiv.org/abs/2112.03227}, 
}

@misc{liu2023liberobenchmarkingknowledgetransfer,
      title={LIBERO: Benchmarking Knowledge Transfer for Lifelong Robot Learning}, 
      author={Bo Liu and Yifeng Zhu and Chongkai Gao and Yihao Feng and Qiang Liu and Yuke Zhu and Peter Stone},
      year={2023},
      eprint={2306.03310},
      archivePrefix={arXiv},
      primaryClass={cs.AI},
      url={https://arxiv.org/abs/2306.03310}, 
}

@misc{zhang2024vlabench,
      title={VLABench: A Large-Scale Benchmark for Language-Conditioned Robotics Manipulation with Long-Horizon Reasoning Tasks}, 
      author={Shiduo Zhang and Zhe Xu and Peiju Liu and Xiaopeng Yu and Yuan Li and Qinghui Gao and Zhaoye Fei and Zhangyue Yin and Zuxuan Wu and Yu-Gang Jiang and Xipeng Qiu},
      year={2024},
      eprint={2412.18194},
      archivePrefix={arXiv},
      primaryClass={cs.RO},
      url={https://arxiv.org/abs/2412.18194}, 
}

@article{li24simpler,
         title={Evaluating Real-World Robot Manipulation Policies in Simulation},
         author={Xuanlin Li and Kyle Hsu and Jiayuan Gu and Karl Pertsch and Oier Mees and Homer Rich Walke and Chuyuan Fu and Ishikaa Lunawat and Isabel Sieh and Sean Kirmani and Sergey Levine and Jiajun Wu and Chelsea Finn and Hao Su and Quan Vuong and Ted Xiao},
         journal = {arXiv preprint arXiv:2405.05941},
         year={2024}
}

@misc{pumacay2024colosseumbenchmarkevaluatinggeneralization,
      title={THE COLOSSEUM: A Benchmark for Evaluating Generalization for Robotic Manipulation}, 
      author={Wilbert Pumacay and Ishika Singh and Jiafei Duan and Ranjay Krishna and Jesse Thomason and Dieter Fox},
      year={2024},
      eprint={2402.08191},
      archivePrefix={arXiv},
      primaryClass={cs.RO},
      url={https://arxiv.org/abs/2402.08191}, 
}

@misc{fang2026generalizableroboticmanipulationdynamic,
      title={Towards Generalizable Robotic Manipulation in Dynamic Environments}, 
      author={Heng Fang and Shangru Li and Shuhan Wang and Xuanyang Xi and Dingkang Liang and Xiang Bai},
      year={2026},
      eprint={2603.15620},
      archivePrefix={arXiv},
      primaryClass={cs.CV},
      url={https://arxiv.org/abs/2603.15620}, 
}

@misc{zhang2026vlm4vlarevisitingvisionlanguagemodelsvisionlanguageaction,
      title={VLM4VLA: Revisiting Vision-Language-Models in Vision-Language-Action Models}, 
      author={Jianke Zhang and Xiaoyu Chen and Qiuyue Wang and Mingsheng Li and Yanjiang Guo and Yucheng Hu and Jiajun Zhang and Shuai Bai and Junyang Lin and Jianyu Chen},
      year={2026},
      eprint={2601.03309},
      archivePrefix={arXiv},
      primaryClass={cs.CV},
      url={https://arxiv.org/abs/2601.03309}, 
}

@misc{hancock2025actionslanguagefinetuningvlms,
      title={Actions as Language: Fine-Tuning VLMs into VLAs Without Catastrophic Forgetting}, 
      author={Asher J. Hancock and Xindi Wu and Lihan Zha and Olga Russakovsky and Anirudha Majumdar},
      year={2025},
      eprint={2509.22195},
      archivePrefix={arXiv},
      primaryClass={cs.RO},
      url={https://arxiv.org/abs/2509.22195}, 
}

@misc{qwenvla,
      title={Qwen-VLA: Unifying Vision-Language-Action Modeling across Tasks, Environments, and Robot Embodiments}, 
      author={Qiuyue Wang and Mingsheng Li and Jian Guan and Jinhui Ye and Sicheng Xie and Yitao Liu and Junhao Chen and Zhixuan Liang and Jie Zhang and Xintong Hu and Xuhong Huang and Pei Lin and Junyang Lin and Dayiheng Liu and Shuai Bai and Jingren Zhou and Jiazhao Zhang and Haoqi Yuan and Gengze Zhou and Hang Yin and Ye Wang and Yiyang Huang and Zixing Lei and Wujian Peng and Delin Chen and Yingming Zheng and Jingyang Fan and Xianwei Zhuang and Xin Zhou and Haoyang Li and Anzhe Chen and Tong Zhang and Xuejing Liu and Yuchong Sun and Ruizhe Chen and Zhaohai Li and Chenxu Lü and Zhibo Yang and Tao Yu and Xionghui Chen},
      year={2026},
      eprint={2605.30280},
      archivePrefix={arXiv},
      primaryClass={cs.RO},
      url={https://arxiv.org/abs/2605.30280}, 
}

@misc{cui2025openhelixshortsurveyempirical,
      title={OpenHelix: A Short Survey, Empirical Analysis, and Open-Source Dual-System VLA Model for Robotic Manipulation}, 
      author={Can Cui and Pengxiang Ding and Wenxuan Song and Shuanghao Bai and Xinyang Tong and Zirui Ge and Runze Suo and Wanqi Zhou and Yang Liu and Bofang Jia and Han Zhao and Siteng Huang and Donglin Wang},
      year={2025},
      eprint={2505.03912},
      archivePrefix={arXiv},
      primaryClass={cs.RO},
      url={https://arxiv.org/abs/2505.03912}, 
}

@misc{brohan2023rt2visionlanguageactionmodelstransfer,
      title={RT-2: Vision-Language-Action Models Transfer Web Knowledge to Robotic Control}, 
      author={Anthony Brohan and Noah Brown and Justice Carbajal and Yevgen Chebotar and Xi Chen and Krzysztof Choromanski and Tianli Ding and Danny Driess and Avinava Dubey and Chelsea Finn and others},
      year={2023},
      eprint={2307.15818},
      archivePrefix={arXiv},
      primaryClass={cs.RO},
      url={https://arxiv.org/abs/2307.15818}, 
}

@misc{kim2024openvlaopensourcevisionlanguageactionmodel,
      title={OpenVLA: An Open-Source Vision-Language-Action Model}, 
      author={Moo Jin Kim and Karl Pertsch and Siddharth Karamcheti and Ted Xiao and Ashwin Balakrishna and Suraj Nair and Rafael Rafailov and Ethan Foster and Grace Lam and Pannag Sanketi and Quan Vuong and Thomas Kollar and Benjamin Burchfiel and Russ Tedrake and Dorsa Sadigh and Sergey Levine and Percy Liang and Chelsea Finn},
      year={2024},
      eprint={2406.09246},
      archivePrefix={arXiv},
      primaryClass={cs.RO},
      url={https://arxiv.org/abs/2406.09246}, 
}

@misc{octomodelteam2024octoopensourcegeneralistrobot,
      title={Octo: An Open-Source Generalist Robot Policy}, 
      author={Octo Model Team and Dibya Ghosh and Homer Walke and Karl Pertsch and Kevin Black and Oier Mees and Sudeep Dasari and Joey Hejna and Tobias Kreiman and Charles Xu and Jianlan Luo and You Liang Tan and Lawrence Yunliang Chen and Pannag Sanketi and Quan Vuong and Ted Xiao and Dorsa Sadigh and Chelsea Finn and Sergey Levine},
      year={2024},
      eprint={2405.12213},
      archivePrefix={arXiv},
      primaryClass={cs.RO},
      url={https://arxiv.org/abs/2405.12213}, 
}

@misc{black2026pi0visionlanguageactionflowmodel,
      title={$\pi_0$: A Vision-Language-Action Flow Model for General Robot Control}, 
      author={Kevin Black and Noah Brown and Danny Driess and Adnan Esmail and Michael Equi and Chelsea Finn and Niccolo Fusai and Lachy Groom and Karol Hausman and Brian Ichter and Szymon Jakubczak and Tim Jones and Liyiming Ke and Sergey Levine and Adrian Li-Bell and Mohith Mothukuri and Suraj Nair and Karl Pertsch and Lucy Xiaoyang Shi and James Tanner and Quan Vuong and Anna Walling and Haohuan Wang and Ury Zhilinsky},
      year={2026},
      eprint={2410.24164},
      archivePrefix={arXiv},
      primaryClass={cs.LG},
      url={https://arxiv.org/abs/2410.24164}, 
}

@inproceedings{gr00tn1_2025,
  archivePrefix = {arxiv},
  eprint     = {2503.14734},
  title      = {{GR00T} {N1}: An Open Foundation Model for Generalist Humanoid Robots},
  author     = {NVIDIA and Johan Bjorck and Fernando Castañeda, Nikita Cherniadev and Xingye Da and Runyu Ding and Linxi "Jim" Fan and Yu Fang and Dieter Fox and Fengyuan Hu and Spencer Huang and Joel Jang and Zhenyu Jiang and Jan Kautz and Kaushil Kundalia and Lawrence Lao and Zhiqi Li and Zongyu Lin and Kevin Lin and Guilin Liu and Edith Llontop and Loic Magne and Ajay Mandlekar and Avnish Narayan and Soroush Nasiriany and Scott Reed and You Liang Tan and Guanzhi Wang and Zu Wang and Jing Wang and Qi Wang and Jiannan Xiang and Yuqi Xie and Yinzhen Xu and Zhenjia Xu and Seonghyeon Ye and Zhiding Yu and Ao Zhang and Hao Zhang and Yizhou Zhao and Ruijie Zheng and Yuke Zhu},
  month      = {March},
  year       = {2025},
  booktitle  = {ArXiv Preprint},
}

@misc{gemini2025robotics,
      title={Gemini Robotics: Bringing AI into the Physical World}, 
      author={Gemini Robotics Team and others},
      year={2025},
      eprint={2503.20020},
      archivePrefix={arXiv},
      primaryClass={cs.RO},
      url={https://arxiv.org/abs/2503.20020},
}

@misc{hao2025mimoembodiedxembodiedfoundationmodel,
      title={MiMo-Embodied: X-Embodied Foundation Model Technical Report}, 
      author={Xiaomi Embodied Intelligence Team},
      year={2025},
      eprint={2511.16518},
      archivePrefix={arXiv},
      primaryClass={cs.RO},
      url={https://arxiv.org/abs/2511.16518},
}

@misc{molmoact2025,
      title={MolmoAct: Action Reasoning Models that can Reason in Space}, 
      author={Jason Lee and Jiafei Duan and Haoquan Fang and Yuquan Deng and Shuo Liu and Boyang Li and Bohan Fang and Jieyu Zhang and Yi Ru Wang and Sangho Lee and Winson Han and Wilbert Pumacay and Angelica Wu and Rose Hendrix and Karen Farley and Eli VanderBilt and Ali Farhadi and Dieter Fox and Ranjay Krishna},
      year={2025},
      eprint={2508.07917},
      archivePrefix={arXiv},
      primaryClass={cs.RO},
      url={https://arxiv.org/abs/2508.07917}
}

@misc{li2026robointerholisticintermediaterepresentation,
      title={RoboInter: A Holistic Intermediate Representation Suite Towards Robotic Manipulation}, 
      author={Hao Li and Ziqin Wang and Zi-han Ding and Shuai Yang and Yilun Chen and Yang Tian and Xiaolin Hu and Tai Wang and Dahua Lin and Feng Zhao and Si Liu and Jiangmiao Pang},
      year={2026},
      eprint={2602.09973},
      archivePrefix={arXiv},
      primaryClass={cs.RO},
      url={https://arxiv.org/abs/2602.09973}, 
}

@misc{zhang2026spatialactionsgroundingvisionlanguageaction,
      title={From Spatial to Actions: Grounding Vision-Language-Action Model in Spatial Foundation Priors}, 
      author={Zhengshen Zhang and Hao Li and Yalun Dai and Zhengbang Zhu and Lei Zhou and Chenchen Liu and Dong Wang and Francis E. H. Tay and Sijin Chen and Ziwei Liu and Yuxiao Liu and Xinghang Li and Pan Zhou},
      year={2026},
      eprint={2510.17439},
      archivePrefix={arXiv},
      primaryClass={cs.RO},
      url={https://arxiv.org/abs/2510.17439}, 
}

@misc{tang2025mindhandpurposefulrobotic,
      title={Mind to Hand: Purposeful Robotic Control via Embodied Reasoning}, 
      author={Peijun Tang and Shangjin Xie and Binyan Sun and Baifu Huang and Kuncheng Luo and Haotian Yang and Weiqi Jin and Jianan Wang},
      year={2025},
      eprint={2512.08580},
      archivePrefix={arXiv},
      primaryClass={cs.RO},
      url={https://arxiv.org/abs/2512.08580}, 
}

@misc{singh2026openaigpt5card,
  title = {{OpenAI GPT-5} System Card},
  author = {Singh, Aaditya and others},
  year = {2026},
  eprint = {2601.03267},
  archivePrefix = {arXiv},
  primaryClass = {cs.CL},
  url = {https://arxiv.org/abs/2601.03267}
}

@misc{gemmateam2025gemma3,
  title = {{Gemma 3} Technical Report},
  author = {{Gemma Team} and others},
  year = {2025},
  eprint = {2503.19786},
  archivePrefix = {arXiv},
  primaryClass = {cs.CL},
  url = {https://arxiv.org/abs/2503.19786}
}

@misc{deitke2024molmopixmoopenweights,
      title={Molmo and PixMo: Open Weights and Open Data for State-of-the-Art Vision-Language Models}, 
      author={Matt Deitke and Christopher Clark and Jordi Salvador and Dustin Schwenk and Aniruddha Kembhavi and Ali Farhadi and others},
      year={2024},
      eprint={2409.17146},
      archivePrefix={arXiv},
      primaryClass={cs.CV},
      url={https://arxiv.org/abs/2409.17146}, 
}

@misc{qwen3.5,
    title  = {{Qwen3.5}: Towards Native Multimodal Agents},
    author = {{Qwen Team}},
    month  = {February},
    year   = {2026},
    url    = {https://qwen.ai/blog?id=qwen3.5}
}

@misc{wu2024deepseekvl2mixtureofexpertsvisionlanguagemodels,
      title={DeepSeek-VL2: Mixture-of-Experts Vision-Language Models for Advanced Multimodal Understanding}, 
      author={Zhiyu Wu and Xiaokang Chen and Zizheng Pan and Xingchao Liu and Wen Liu and Damai Dai and Huazuo Gao and Yiyang Ma and Chengyue Wu and Bingxuan Wang and Zhenda Xie and Yu Wu and Kai Hu and Jiawei Wang and Yaofeng Sun and Yukun Li and Yishi Piao and Kang Guan and Aixin Liu and Xin Xie and Yuxiang You and Kai Dong and Xingkai Yu and Haowei Zhang and Liang Zhao and Yisong Wang and Chong Ruan},
      year={2024},
      eprint={2412.10302},
      archivePrefix={arXiv},
      primaryClass={cs.CV},
      url={https://arxiv.org/abs/2412.10302}, 
}

@misc{khazatsky2025droidlargescaleinthewildrobot,
      title={DROID: A Large-Scale In-The-Wild Robot Manipulation Dataset}, 
      author={Alexander Khazatsky and Karl Pertsch and Suraj Nair and Ashwin Balakrishna and Sudeep Dasari and Siddharth Karamcheti and Soroush Nasiriany and Mohan Kumar Srirama and Lawrence Yunliang Chen and Kirsty Ellis and others},
      year={2025},
      eprint={2403.12945},
      archivePrefix={arXiv},
      primaryClass={cs.RO},
      url={https://arxiv.org/abs/2403.12945}, 
}

@misc{comanici2025gemini25pushingfrontier,
      title={Gemini 2.5: Pushing the Frontier with Advanced Reasoning, Multimodality, Long Context, and Next Generation Agentic Capabilities},
      author={Gheorghe Comanici and others},
      year={2025},
      eprint={2507.06261},
      archivePrefix={arXiv},
      primaryClass={cs.CL},
      url={https://arxiv.org/abs/2507.06261},
}

@misc{anthropic2026fable51card,
  author       = {{Anthropic}},
  title        = {System Card: Claude Fable 5.1 \& Claude Mythos 5.1},
  year         = {2026},
  url = {https://www.anthropic.com/claude-fable-5-1-mythos-5-1-system-card}
}

@misc{fu2026capxframeworkbenchmarkingimproving,
      title={CaP-X: A Framework for Benchmarking and Improving Coding Agents for Robot Manipulation}, 
      author={Letian Fu and Justin Yu and Karim El-Refai and Ethan Kou and Haoru Xue and Huang Huang and Wenli Xiao and Guanzhi Wang and Dantong Niu and Fei-Fei Li and Guanya Shi and Jiajun Wu and Shankar Sastry and Yuke Zhu and Ken Goldberg and Linxi "Jim" Fan},
      year={2026},
      eprint={2603.22435},
      archivePrefix={arXiv},
      primaryClass={cs.RO},
      url={https://arxiv.org/abs/2603.22435}, 
}
